\documentclass[preprint,authoryear,12pt]{elsarticle}

\usepackage{graphicx}
\usepackage{mathtools}
\usepackage{booktabs}
\usepackage{multirow}
\usepackage{microtype}

\usepackage{rotating}
\usepackage{hyperref}
\usepackage{amssymb}
\usepackage{afterpage}
\usepackage{boldcommands}

\usepackage{xcolor}
\usepackage{changebar}
\usepackage{marginnote}
\usepackage{tikz}
\usepackage{etoolbox}
\usepackage{caption}
\usepackage{pgfplots}
\usetikzlibrary{spy,calc}
\usepackage[framemethod=TikZ]{mdframed}
\usetikzlibrary{trees,shapes,snakes}
\usetikzlibrary{plotmarks}
\usetikzlibrary{positioning}
\usetikzlibrary{backgrounds}
\usetikzlibrary{fit,calc,decorations.pathreplacing,matrix}
\usetikzlibrary{decorations.text}
\usetikzlibrary{arrows}
\usepackage{fullpage}
\usepackage{rotating}

\usepackage{tikz}
\usepackage{pgfplots}

\usetikzlibrary{positioning}

\newcommand{\0}{\hspace*{1.1ex}}

\tikzstyle{na} = [baseline=-.5ex]

\journal{NeuroImage}

\begin{document}

	\begin{frontmatter}

		\title{PSACNN: Pulse Sequence Adaptive Fast Whole Brain Segmentation}

		\author[mgh,hms]{Amod Jog}
		\ead[mgh]{ajog@mgh.harvard.edu}
		\author[mgh]{Andrew Hoopes}
		\author[mgh,hms]{Douglas N. Greve}
		\author[mgh,dtu]{Koen Van Leemput}
		\author[mgh,hms,hst]{Bruce Fischl}

		\address[mgh]{Athinoula A. Martinos Center for Biomedical Imaging, Department of Radiology, Massachusetts General Hospital, Charlestown, MA 02129}
		\address[dtu]{Department of Health Technology, Technical University of Denmark}
		\address[hms]{Department of Radiology, Harvard Medical School, United States}
		\address[hst]{Division of Health Sciences and Technology and Engineering and Computer Science MIT, Cambridge, MA, United States}

\begin{abstract} 
With the advent of convolutional neural networks~(CNN), supervised learning
methods are increasingly being used for whole brain segmentation. However, a large, manually annotated training dataset
of labeled brain images required to train such supervised methods is frequently difficult to obtain or create. In addition, existing training datasets are generally acquired with a homogeneous
magnetic resonance imaging~(MRI) acquisition protocol. CNNs trained on such datasets are unable to generalize on test data with
different acquisition protocols. Modern neuroimaging studies and clinical trials are necessarily
multi-center initiatives with a wide variety of acquisition protocols. Despite
stringent protocol harmonization practices, it is very difficult to standardize
the gamut of MRI imaging parameters across scanners, field strengths,
receive coils etc., that affect image contrast. In this paper we propose a
CNN-based segmentation algorithm that, in addition to being highly accurate and
fast, is also resilient to variation in the input acquisition.
Our approach relies on building approximate forward models of pulse sequences
that produce a typical test image. For a given pulse sequence, we use its forward model to generate plausible, synthetic training examples that appear as if they were acquired in a scanner with that pulse sequence. Sampling over a
wide variety of pulse sequences results in a wide variety of augmented training examples that help build an image contrast invariant model. Our method trains a single CNN that can segment input MRI images with acquisition parameters as disparate as $T_1$-weighted and $T_2$-weighted contrasts with only $T_1$-weighted training data. The segmentations generated are highly accurate with state-of-the-art results~(overall Dice overlap$=0.94$), with a fast run time~($\approx$ 45 seconds), and consistent across a wide range of acquisition protocols.
\end{abstract}


\end{frontmatter}

\section{Introduction}
Whole brain segmentation is one of the most important tasks in a neuroimage processing pipeline, and is a well-studied problem~\citep{fischl2004aseg, shiee2010ni,wang2013malf,asman2012tmi,wachinger2017}. The segmentation task
involves taking as input an MRI image, usually a $T_1$-weighted~($T_1$-w) image, of the head, and generating a labeled image with each voxel getting a label of the structure it lies in. Most segmentation algorithms output labels
for right$/$left white matter, gray matter cortex, and subcortical structures such as the thalamus, hippocampus, amygdala, and others. Structure volumes and shapes estimated from a volumetric segmentation are routinely used as biomarkers to quantify differences between healthy and diseased
populations~\citep{fischl2004aseg}, track disease progression~\citep{frisoni2009,jack2011,davatzikos2005}, and study aging~\citep{Resnick2000}. Segmentation of a $T_1$-w image is also used
to localize and analyze the functional and diffusion signals from labeled
anatomy after modalities such as functional MRI~(fMRI), PET~(positron emission tomography), and diffusion MRI are co-registered to the $T_1$-w image. This enables the study of effects of external intervention or treatment on various properties of the segmented neuroanatomy~\citep{desbordes2012,mattson2014}.

An ideal segmentation algorithm needs to be highly accurate. Accuracy is frequently measured with metrics such as Dice coefficient and Jaccard index,
that quantify overlap of predicted labels with a manually labeled ground truth. An accurate segmentation algorithm is more sensitive to and can detect the
subtle changes in volume and shape of brain structures in patients with neurodegenerative diseases~\citep{jack2011}.
Fast processing time is also a
desirable property for segmentation algorithms. Segmentation is one of the
time-consuming bottlenecks in neuroimaging pipelines~\citep{fischl2004aseg}.
A fast, memory-light segmentation algorithm can process a larger dataset quickly
and can potentially be adopted in real-time image analysis pipelines.

For MRI, however, it is also critical that the segmentation algorithm be robust to variations in the contrast properties of the input images. MRI is an extremely versatile modality, and
a variety of image contrasts can be achieved by changing the multitude of imaging parameters.
Large, modern MRI studies acquire imaging data simultaneously
at multiple centers across the globe to gain access to a larger, more diverse pool of subjects~\citep{enigma, adni}. It is very difficult
to perfectly harmonize scanner manufacturers, field strengths, receive coils, pulse
sequences, resolutions, scanner software versions, and many other imaging parameters across different
sites. Variation in imaging parameters leads to variation in intensities
and image contrast across sites, which in turn leads to variation in segmentation results. Most segmentation algorithms are not designed to be robust to these variations~\citep{han2006ni,jovicich2013ni,nyul1999mrm,focke2011}. Segmentation algorithms can introduce a site-specific bias in any analysis of multi-site data that is dependent on structure volumes or shapes obtained from that algorithm. In most analyses, the inter-site and inter-scanner variations are regressed out as
confounding variables~\citep{kostro2014} or the MRI images are pre-processed
in order to provide consistent input to the segmentation algorithm~\citep{nyul1999mrm,madabhushi2006mp,roy2013tmi,jog2015media}.
However, a segmentation algorithm that is robust to scanner variation precludes the need to pre-process the inputs.

\subsection{Prior Work}
Existing whole brain segmentation algorithms can be broadly classified into three types:
(a)~model-based, (b)~multi-atlas registration-based, and (c)~supervised learning-based. Model-based algorithms~\citep{fischl2004aseg,pohl2006,pizer2003,patenaude2011,puonti2016samseg}
fit an underlying statistical atlas-based model to the observed intensities of the input image and perform maximum a posteriori~(MAP) labeling. They
assume a functional form~(e.g. Gaussian) of intensity distributions and results
can degrade if the input distribution differs from this assumption. Recent work by~\cite{puonti2016samseg} builds a parametric generative model for
brain segmentation that leverages manually labeled training data and also adapts to changes in input image contrast. In a similar vein, work
by~\cite{bazin2008media} also uses a statistical atlas to minimize an energy functional based on fuzzy c-means clustering of intensities. Model-based methods are usually computationally intensive, taking 0.5--4 hours to run. With the exception of the work by~\cite{puonti2016samseg} and ~\cite{bazin2008media}, most model-based methods are not explicitly designed to be robust to variations in input $T_1$-w contrast and fail to work on input images acquired with a completely different contrast such a proton density weighted~($P_D$-w) or $T_2$-weighted~($T_2$-w).

Multi-atlas registration and label fusion~(MALF) methods~\citep{rohlfing2004,heckemann2006,sabuncu2010tmi,zikic2014,asman2012tmi,wang2013malf,wu2014} use an atlas set of images that comprises pairs of intensity images and their corresponding manually labeled images. The atlas
intensity images are deformably registered to the input intensity image, and
the corresponding atlas label images are warped using the estimated
transforms into the input image space. For each voxel in the input image space,
a label is estimated using a label fusion approach. MALF algorithms achieve state-of-the-art segmentation accuracy
for whole brain segmentation~\citep{asman2012tmi,wang2013malf} and have previously won multi-label segmentation challenges~\citep{landman2012}.
However, they require multiple computationally expensive registrations,
followed by label fusion. Efforts have been made to reduce the computational cost of deformable registrations with linear registrations~\citep{coupe2011}, or reducing the task to a single deformable
registration~\citep{zikic2014}. However, if the input image contrast is significantly different from the atlas intensity images, registration quality
becomes inconsistent, especially across different multi-scanner datasets.  If the input image contrast is completely different~($P_D$-w or $T_2$-w) from the atlas images~(usually $T_1$-w), the registration algorithm needs to use a
modality independent optimization metric such as cross-correlation
or mutual information, which also affects registration consistency across
datasets.

Accurate whole brain segmentation using supervised machine learning
approaches has been made possible in the last few years by the success of
deep learning methods in medical imaging. Supervised segmentation approaches built on CNNs have produced accurate segmentations with a runtime of a few seconds or minutes~\citep{niftynet17,wachinger2017,guharoy2018}. These methods
have used 2D slices~\citep{guharoy2018} or 3D patches~\citep{niftynet17,wachinger2017} as inputs to custom-built fully convolutional network architectures, the U-Net~\citep{unet2015} or residual architectures~\citep{he2016} to produce voxel-wise semantic segmentation
outputs. The training of these deep networks can take hours to days, but the
prediction time on an input image is in the range of seconds to minutes, thus
making them an attractive proposition for whole brain segmentation. CNN  architectures for whole brain segmentation tend to have millions of trainable parameters
but the training data is generally limited. A manually labeled training
dataset can consist of only $20$--$40$ labeled images~\citep{buckner2004protocol,landman2012} due to the labor-intensive
nature of manual annotation. Furthermore, the training
dataset is usually acquired with a fixed acquisition protocol, on the same
scanner, within a short period of time, and generally without any
substantial artifacts, resulting in a pristine set of training images.
Unfortunately, a CNN trained on such limited data
is unable to generalize to an input test image acquired with a different
acquisition protocol and can easily overfit to the training acquisitions.
Despite the powerful local and global context that CNNs generally provide,
they are more vulnerable to subtle contrast differences between training
and test MRI images than model-based and MALF-based methods. A recent work
by~\cite{karani2018lifelong}
describes a brain segmentation CNN trained to consistently segment multi-scanner and multi-protocol data. This framework adds
a new set of batch normalization parameters in the network as it encounters training data from a new acquisition protocol. To learn these parameters,
the framework requires that a small number of images from the new protocol be manually labeled, which may not always be possible.


\subsection{Our Contribution}
\label{sec:contribution}
We describe Pulse Sequence Adaptive Convolutional Neural Network (PSACNN); a CNN-based whole brain MRI segmentation approach with an
augmentation scheme that models the forward models of pulse sequences to
generate a wide range of synthetic training examples during the training
process.
Data augmentation is regularly used in training of CNNs when training data is
scarce. Typical augmentation in computer vision applications involves
rotating, deforming, and adding noise to an existing training example, pairing
it with a suitably transformed training label, and adding it to the training
batch data. Once trained this way, the CNN is expected to be robust to these
variations, which were not originally present in the given training dataset.

In PSACNN, we aim to augment CNN training by using existing training
images and creating versions of them with a different contrast, for a wide
range of contrasts. The CNN thus trained will be robust to contrast variations.
We create synthetic versions of training images such that they appear to have been scanned with different pulse sequences for a range of pulse sequence parameters. These synthetic images are created by applying pulse sequence
imaging equations to the training tissue parameter maps. We formulate simple approximations of the highly nonlinear pulse sequence imaging equations for a set of commonly used pulse sequences. Our approximate imaging equations typically have a much smaller number~(3 to be precise) of imaging parameters.
Estimating these parameters would typically need multi-contrast images that are
not usually available, so we have developed a procedure to estimate these approximate imaging parameters using intensity information present in that image alone.

When applying an approximate imaging equation to tissue parameter maps, changing the imaging parameters of the approximate imaging equation changes the contrast of the generated synthetic image. In PSACNN, we uniformly sample over the approximate pulse sequence parameter space to produce candidate pulse sequence imaging equations and apply them to
the training tissue parameters to synthesize images of training brains with
a wide range of image contrasts. Training using these synthetic images results
in a segmentation CNN that is robust to the variations in image contrast.

A previous version of PSACNN was published as a conference
paper~\citep{jog2018miccai}. In this paper, we have extended it by removing
the dependence on the FreeSurfer atlas registration for features. We have also
added pulse sequence augmentation for $T_2$-weighted sequences, thus enabling
the same trained CNN to segment $T_1$-w and $T_2$-w images without an
explicitly $T_2$-w training dataset. We have also included additional
segmentation consistency experiments conducted on more diverse multi-scanner
datasets.

Our paper is structured as follows: In Section~\ref{sec:method}, we provide a background on MRI image formation, followed by a description of
our pulse sequence forward models and the PSACNN training workflow. In
Section~\ref{sec:results}, we show an evaluation of PSACNN performance on a
wide variety of multi-scanner datasets and compare it with state-of-the-art
segmentation methods. Finally, in Section~\ref{sec:conclusion} we briefly summarize our observations and outline directions for future work.

\section{Method}
\label{sec:method}
\subsection{Background: MRI Image Formation}
\label{sec:background}
In MRI the changing electromagnetic field stimulates tissue inside a voxel
to produce the voxel intensity signal. Let $S$ be the acquired MRI magnitude
image. $S(\boldx)$, the intensity at voxel $\boldx$ in the brain is a function of (a) intrinsic tissue parameters and (b) extrinsic imaging
parameters. We will refer to the tissue parameters as nuclear magnetic resonance or NMR parameters in order to distinguish them from the imaging
parameters. The principal NMR parameters include proton density~($\rho(\boldx)$), longitudinal~($T_1(\boldx)$) and transverse~($T_2(\boldx)$), and $T^{*}_{2}(\boldx)$ relaxation times. In this work we only focus on $\rho$, $T_1$, $T_2$ and denote them together as $\boldbeta(\boldx) = [\rho(\boldx), T_1(\boldx), T_2(\boldx)]$.
Let the image $S$ be acquired using a pulse sequence $\Gamma_S$. Commonly used
pulse sequences include Magnetization Prepared Gradient Echo~(MPRAGE)~\citep{mugler1990mrm}, Multi-Echo MPRAGE~\citep{vanderkouve2008memprage} Fast Low Angle Shot~(FLASH),
Spoiled Gradient Echo~(SPGR), T2-Sampling Perfection with Application optimized
Contrasts using different flip angle Evolution~(T2-SPACE)~\citep{mugler2014t2space} and others.
Depending on the type of the pulse sequence $\Gamma_S$, its extrinsic
imaging parameters can include repetition time~($TR$), echo time~($TE$),
flip angle~($\alpha$), gain~($G$), and many others. The set of imaging
parameters is denoted by $\boldTheta_S = \{TR, TE, \alpha, G, \ldots\}$.
The relationship between voxel intensity $S(\boldx)$ and the NMR parameters~($\boldbeta(\boldx)$) and imaging parameters~($\boldTheta_S$) is
encapsulated by the imaging equation as shown in Eqn.~\ref{eq:imgeqn}:
\begin{equation}
S(\boldx) = \Gamma_S(\boldbeta(\boldx); \boldTheta_S).
\label{eq:imgeqn}
\end{equation}
For example, for the FLASH sequence, the imaging parameters are
$\boldTheta_{FLASH} = \{ TR, TE, \alpha, G\}$, and the imaging equation is shown in Eqn.~\ref{eq:flash}~\citep{handbookBernstein}:
\begin{equation}
\Gamma_{\textrm{FLASH}}(\boldbeta(\boldx); \boldTheta_{FLASH}) = G{\rho(\boldx)} \sin \alpha \frac{(1 - e^{-\frac{TR}{T_1(\boldx)}})}{1 - \cos\alpha e^{-\frac{TR}{T_1(\boldx)}}}e^{-\frac{TE}{T_{2}^{*}(\boldx)}}.
\label{eq:flash}
\end{equation}

Most pulse sequences do not have a closed form imaging equation and a Bloch simulation is needed to estimate the voxel intensity~\citep{handbookBernstein}. There has been some work in
deriving the imaging equation for MPRAGE~\citep{Deichmann2000,wang2014mprage}.
Unfortunately, these closed form equations do not always match the scanner implementation of the pulse sequence. Different scanner manufacturers and
scanner software versions have different implementations of the same pulse
sequence. Sometimes additional pulses like fat suppression are
added to improve image quality, further deviating from the closed form theoretical equation. Therefore, if such a closed form theoretical imaging
equation with known imaging parameters~(from the image header) is applied to
NMR maps~($[\rho, T_1,T_2]$) of a subject, the resulting synthetic image will
systematically differ from the observed, scanner-acquired image. To mitigate
this difference, we deem it essential to estimate the imaging parameters purely from the observed image intensities themselves. This way, when the imaging
equation with the estimated parameters is applied to the NMR maps, we are
likely to get a synthetic image that is close to the observed, scanner-acquired
image. However, to enable estimation of imaging parameters purely from the
image intensities requires us to approximate the closed form theoretical
imaging equation to a much simpler form and with fewer parameters.
Therefore, we do not expect the estimated parameters of an approximate imaging
equation to match with the scanner parameters (such as $TR$ or $TE$). The
approximation of the imaging equation and its estimated parameters are deemed
satisfactory if they produce intensities similar to an observed,
scanner-acquired image, when applied to NMR maps.
Our approximations of the theoretical closed form imaging equations for a
number of commonly used pulse sequences are described next in
Section~\ref{sec:estimation}.


\subsection{Pulse Sequence Approximation and Parameter Estimation}
\label{sec:estimation}
\begin{figure}[!ht] \tabcolsep 1pt
	\begin{center}

	\begin{tabular}{ccc}
	\includegraphics[width=0.33\textwidth]{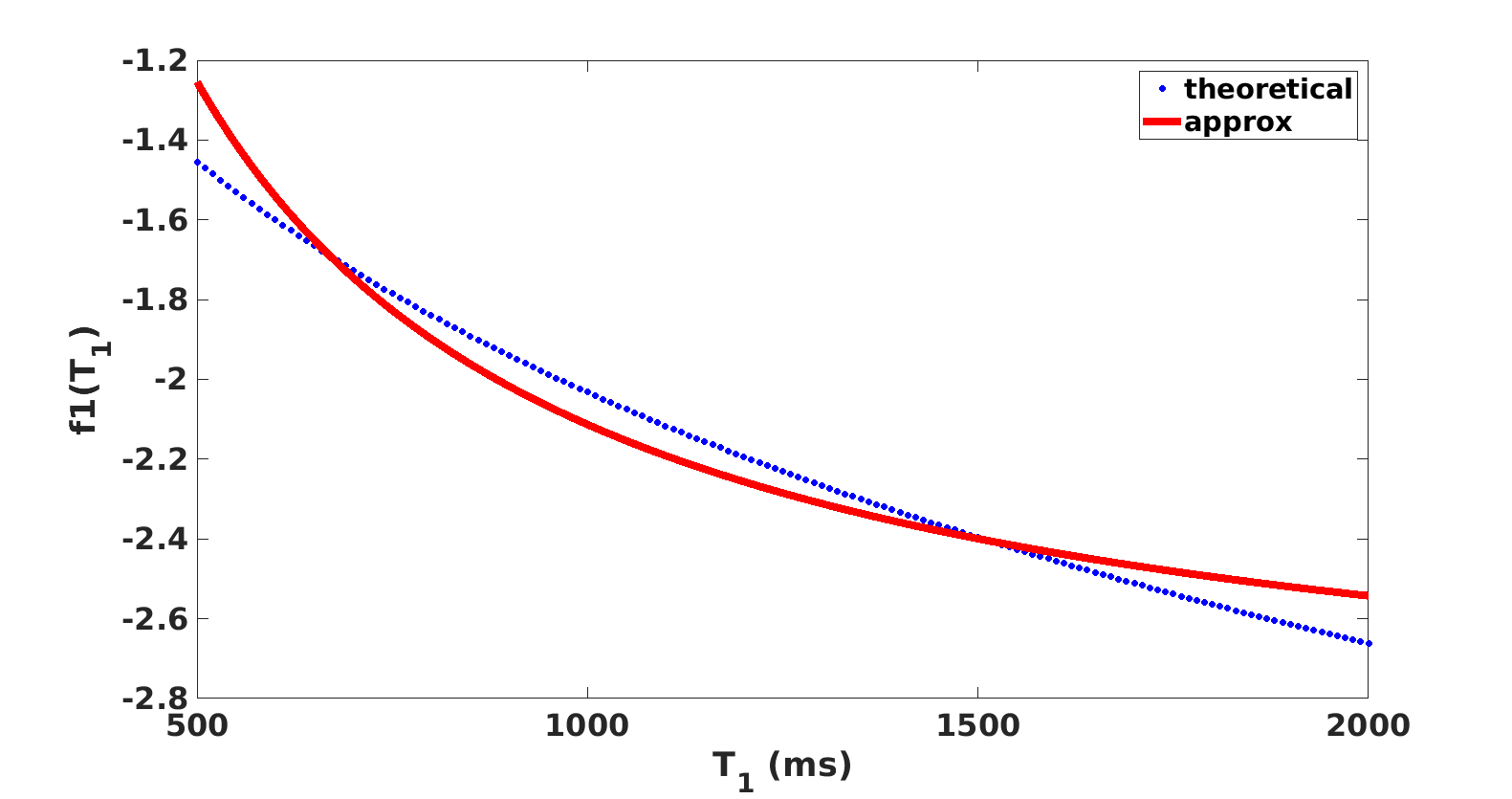} &
	\includegraphics[width=0.33\textwidth]{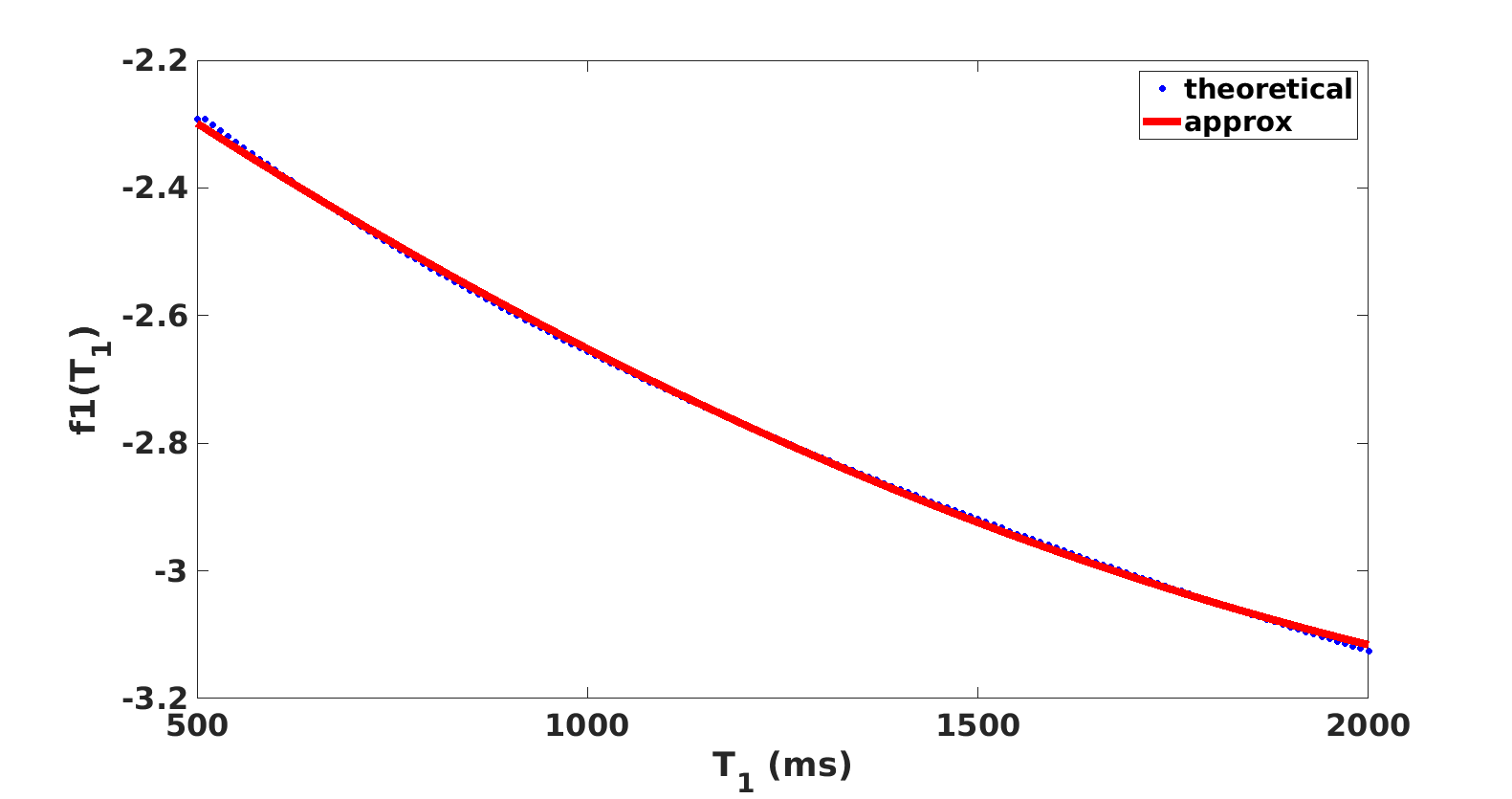} &
	\includegraphics[width=0.33\textwidth]{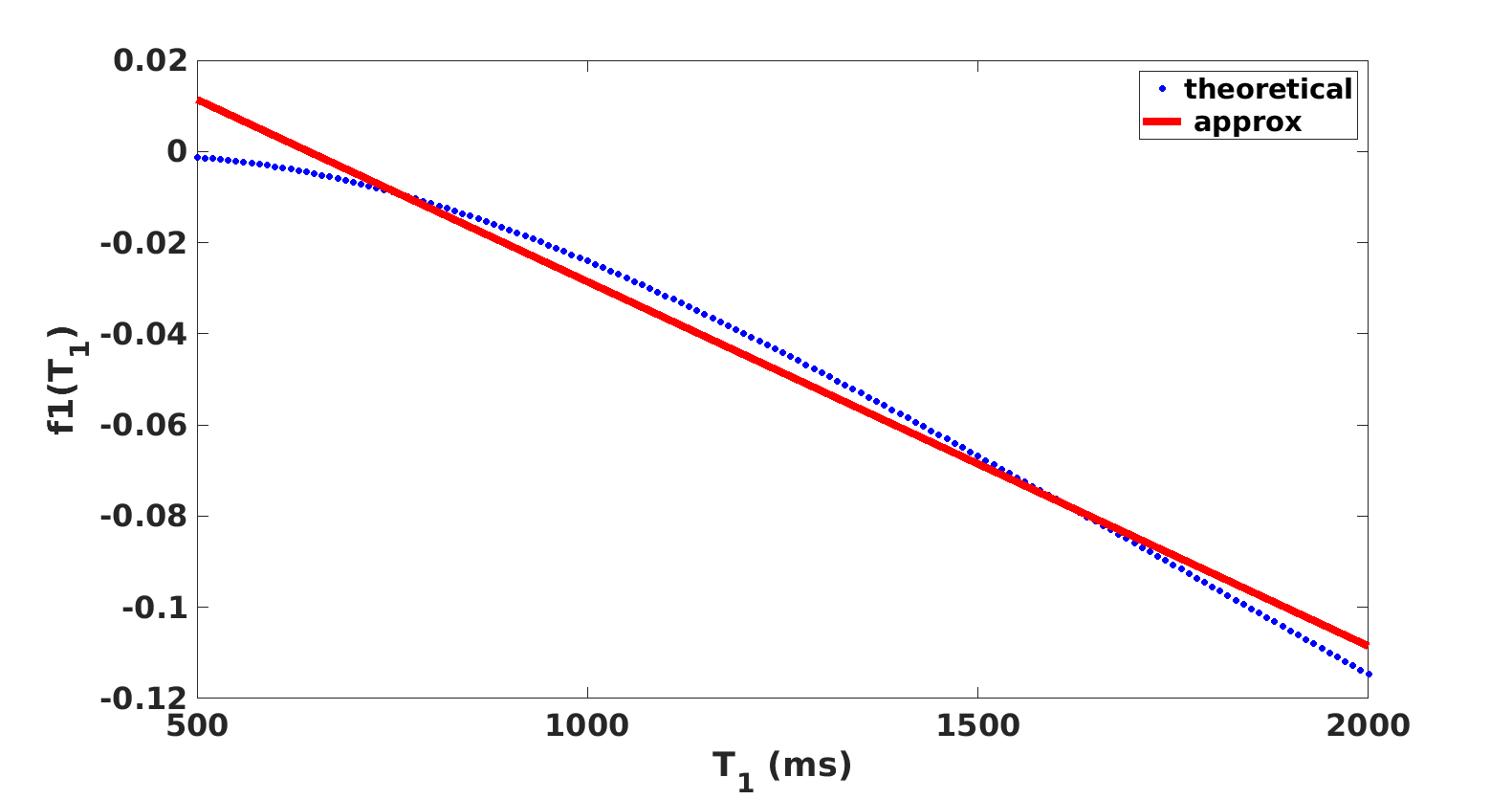}
	\\
		(a) & (b) & (c)
		\end{tabular}
\end{center}
	\caption{(a) Fit of $T_1$ component in the FLASH equation~(blue) and our approximation~(red), (b) fit of $T_1$ component~(blue) with our approximation~(red) for MPRAGE, (c) fit of $T_1$ component~(blue)
	with our approximation~(red) for T2-SPACE}
	\label{fig:approx}
\end{figure}

In this work we primarily work with FLASH, MPRAGE, SPGR, and T2-SPACE
images. Our goal is to be able to estimate the pulse sequence parameters $\boldTheta_S$ from the intensity information of $S$. Intensity histograms of most pulse sequence images present peaks for the three main tissue classes in the brain: cerebro-spinal fluid~(CSF), gray matter~(GM), and white matter~(WM). The mean class intensities can be robustly estimated by fitting a three-class Gaussian mixture model to the intensity histogram. Let $S_c$, $S_g$, $S_w$ be the mean intensities of CSF, GM, and WM, respectively. Let $\boldbeta_{c}$, $\boldbeta_{g}$,
$\boldbeta_{w}$ be the mean NMR $[\rho, T_1, T_2]$ values for CSF, GM, and WM
classes, respectively. In this work, we extracted these values from a previously acquired dataset with relaxometry acquisitions~\citep{Fischl2004}, but they could also have been
obtained from previous studies~\citep{Wansapura1999}. We assume that applying
the pulse sequence equation $\Gamma_S$ to the mean class
NMR values results in mean class intensities in $S$. This assumption leads to a
three equation system as shown in Eqns.~\eqref{eq:sc}--\eqref{eq:sw}, where the
unknown is $\boldTheta_S$:
\begin{align}
S_c & =  \Gamma_S(\boldbeta_{c}; \boldTheta_S), \label{eq:sc}\\
S_g & =  \Gamma_S(\boldbeta_{g}; \boldTheta_S), \label{eq:sg}\\
S_w & =  \Gamma_S(\boldbeta_{w}; \boldTheta_S). \label{eq:sw}
\end{align}
To solve a three equation system, it is necessary that $\boldTheta_S$
have three or fewer unknown parameters. From the FLASH imaging equation in Eqn.~\ref{eq:flash}, it is clear that there are more than three unknowns~(four, to be precise) in the closed form imaging equation, which is also true for most pulse sequences. Therefore, to enable
us to solve the system in Eqns.~\eqref{eq:sc}--\eqref{eq:sw}, we formulate
a three-parameter approximation for pulse sequences like FLASH, SPGR, MPRAGE,
and T2-SPACE. Our approximation for the FLASH sequence is shown in Eqns.~\eqref{eq:flash_approx1}--\eqref{eq:flash_approx2}:
 \begin{align}
 \log(S_{\textrm{FLASH}}) & = \log(G\sin\alpha) + \log({\rho}) + \log(\frac{(1 - e^{-\frac{TR}{T_1}})}{1 - \cos\alpha e^{-\frac{TR}{T_1}}}) -\frac{TE}{T_{2}^{*}}, \label{eq:flash_approx1}  \\
 & = \theta_0 + \log(\rho) + f1(T_1) + f2(T_2), \label{eq:flash_approx1p5}\\
 &\approx \theta_0 + \log(\rho) + \frac{\theta_1}{T_1} + \frac{\theta_2}{T_2},
 \label{eq:flash_approx2}
 \end{align}
 where $\boldTheta_{\textrm{FLASH}} = \{\theta_0, \theta_1, \theta_2\}$ forms
 our parameter set.
We replace ${T_2}^*$ by $T_2$ as they are highly correlated quantities due to
the relation $1/{T^{*}_{2}} = 1/T_2 + 1/{T_2^{'}}$,
where $1/{T_2^{'}}$ is dependent on field inhomegeneities~\citep{chavhan2009radiographics} that are assumed to have been
eliminated by an inhomogeneity correction algorithm in
pre-processing~\citep{N3}.
$f1(T_1)$ and $f2(T_2)$ in Eqn.~\ref{eq:flash_approx1p5} model the contribution
of $T_1$ and $T_2$ components respectively to the voxel intensity.
Figure~\ref{fig:approx}(a) shows the theoretical $f1(T_1)$ and our approximation
$\theta_1/{T_1}$ for FLASH. For the range of values of $T_1$ in the human
brain $(500,3000)$~ms at 1.5~T, our approximation fits $f1(T_1)$ well.
The error increases for long $T_1$ regions, which are generally CSF voxels in
the brain, and relatively easy to segment.
The SPGR pulse sequence is very similar to FLASH and so we use the same
approximation for it~\citep{handbookBernstein}.

For the MPRAGE sequence, we
chose to approximate a theoretical derivation provided by~\cite{wang2014mprage}.
Assuming the echo time $TE$ is small, which it typically is to avoid
${T_2}^*$-induced loss, the ${T_2}^*$ effects in our approximation are not
significant. Therefore, we use two parameters $\theta_1$ and $\theta_2$ to
model a quadratic dependence on $T_1$, which is an approximation of the
derivation by~\cite{wang2014mprage}, resulting in Eqn.~\eqref{eq:mprage_approx}
as our three-parameter approximation for MPRAGE. Comparison with the
theoretical dependence on $T_1$ is shown in Fig.~\ref{fig:approx}(b), and our
approximation shows a very good fit.
\begin{equation}
\log(S_{\textrm{MPRAGE}}) \approx \theta_0 + \log(\rho) + \theta_1T_1 + \theta_2{T_1}^2.
\label{eq:mprage_approx}
\end{equation}

T2-SPACE is a turbo spin echo sequence. The intensity produced at the $n^{th}$
echo of a turbo spin echo sequence is given by Eqn.~\ref{eq:t2space}:
\begin{equation}
\log(S_{T2-SPACE}) = \log(G) + \log(\rho) + \log(1 - Fe^{-\frac{TD}{T_1}}) - \frac{TE_{n}}{T_2},
\label{eq:t2space}
\end{equation}
where $F\approx1$ for a first order approximation~\citep{handbookBernstein},
$TD$ is the time interval between last echo and $TR$, and $TE_n$ is the
$n^{th}$ echo time. Our three-parameter approximation for Eqn.~\ref{eq:t2space}
is given in Eqn.~\ref{eq:t2space_approx}:
\begin{equation}
\log(S_{T2SPACE}) \approx \theta_0 + \log(\rho) + \theta_1 T_1 + \frac{\theta_2}{T_2}.
\label{eq:t2space_approx}
\end{equation}
The comparison with the theoretical equation as shown in Fig.~\ref{fig:approx}(c)
shows good fit for $T_1$ dependence for $T_1$ values in the GM-WM range. The error increases for very short $T_1$~(bone) and very long $T_1$~(CSF) voxels.
This has minimal impact on the segmentation accuracy as they have sufficient
contrast to segment regardless of small deviations from the model.

Our three-parameter approximations are such that the equation system in Eqns.~\eqref{eq:sc}--\eqref{eq:sw} becomes linear in terms of $\boldTheta_S$
and can be easily and robustly solved for each of the pulse sequences. The
exact theoretical equations have more than three parameters, and even if they
had fewer parameters, the exponential interaction of parameters with $T_1$ and
$T_2$ renders solving the highly nonlinear system very unstable. Using our
formulated approximate versions of pulse sequence equations, given any image
$S$ and the pulse sequence used to acquire it, we can estimate the three
parameter set $\hat{\boldTheta}_{S}$, denoted by hat to distinguish it from
the unknown parameter set $\boldTheta_S$.

\begin{figure}[!ht] \tabcolsep 1pt
	\centerline{
		\begin{tabular}{ccc}
			\includegraphics[width=.33\textwidth]{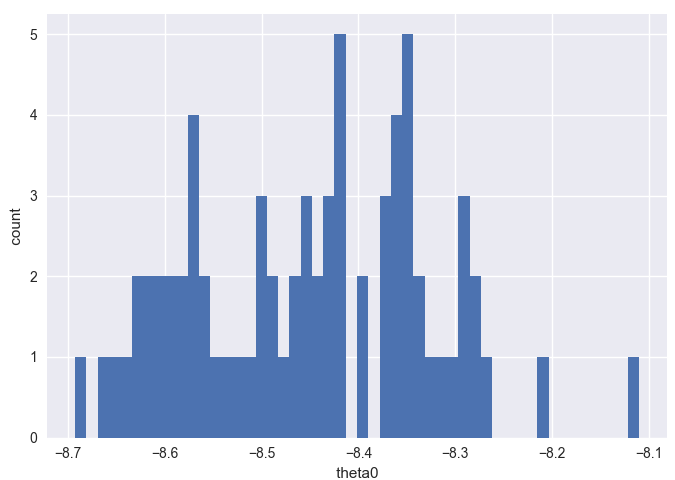} &
			\includegraphics[width=.33\textwidth]{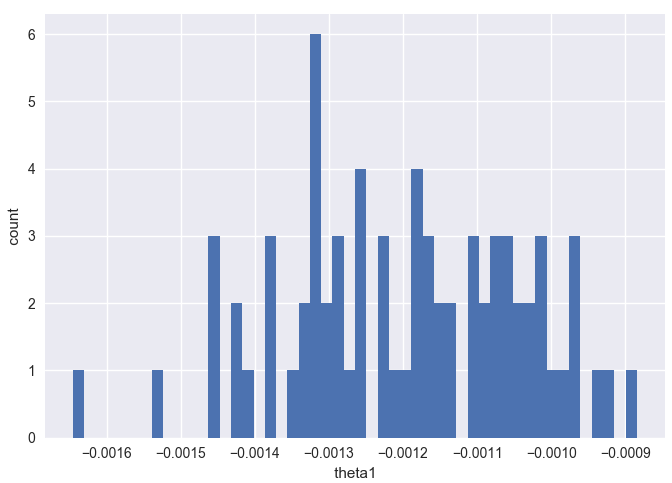} &
			\includegraphics[width=.33\textwidth]{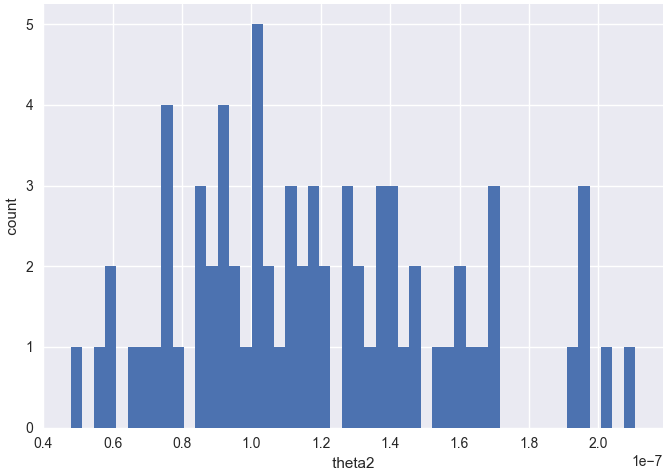} \\
			(a) $\theta_0$ distribution & (b) $\theta_1$ distribution & (c) $\theta_2$ distribution
	\end{tabular}
		}
		\caption{Estimated MPRAGE parameters $\theta_0$, $\theta_1$, and $\theta_2$
		distributions for a variety of 1.5~T and 3~T datasets from Siemens Trio/SONATA/Prisma/Avanto, GE Signa scanners.}
		\label{fig:thetaspace}
	\end{figure}
More importantly, the three-parameter $[\theta_0, \theta_1, \theta_2]$ approximations establish a compact, 3-dimensional parameter space for a pulse sequence. Given any test image, its approximate pulse sequence parameter set is a point in this 3-dimesional parameter space, found by calculating the mean class intensities and solving the equation system in Eqns.~\eqref{eq:sc}--\eqref{eq:sw}. We estimated the parameter sets for a variety
of test MPRAGE images acquired on 1.5~T and 3~T scanners from Siemens Trio/Vision/SONATA/Avanto/Prisma and GE Signa scanners using our method. The estimated parameter distributions are shown in Figs.~\ref{fig:thetaspace}(a)--~\ref{fig:thetaspace}(c). These give us an idea of the range of parameter
values and their distribution for the MPRAGE sequence. We use this information to grid the parameter space in 50 bins between [0.8$\min(\theta_m)$, 1.2$\max(\theta_m)$] for $m \in \{0, 1, 2\}$ such that any extant test dataset parameters are included in this interval and are close to a grid point. We create such a 3D grid for FLASH/SPGR and T2-SPACE as well. In Section~\ref{sec:training} we describe PSACNN training that depends on sampling this grid to come up with candidate pulse
sequence parameter sets. We will use these candidate parameter sets to create augmented, synthetic training images.

\subsection{PSACNN Training}
\label{sec:training}
\begin{figure}[!ht] \tabcolsep 1pt
	\centerline{
		\begin{tabular}{ccccc}
			\includegraphics[width=.2\textwidth]{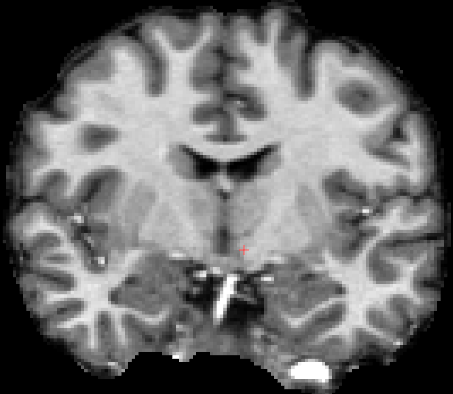} &
			\includegraphics[width=.2\textwidth]{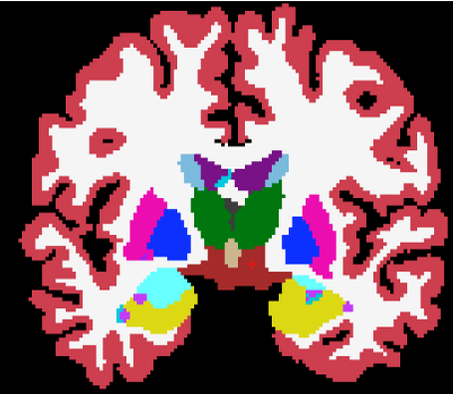} &
			\includegraphics[width=.2\textwidth]{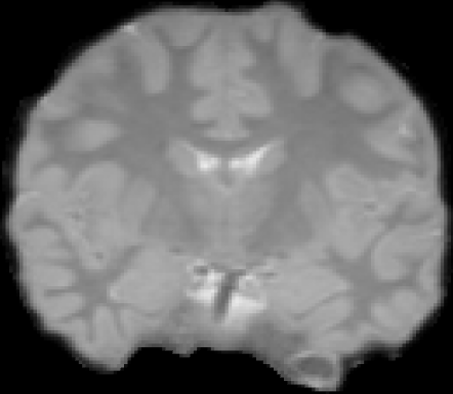} &
			\includegraphics[width=.2\textwidth]{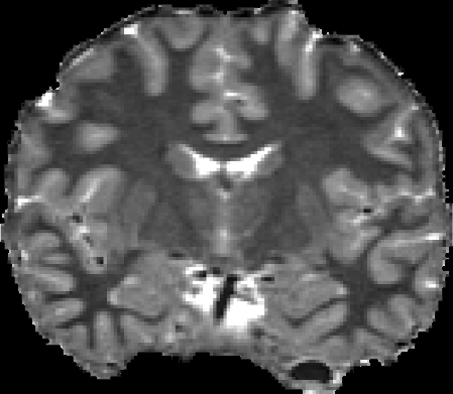}&
			\includegraphics[width=.2\textwidth]{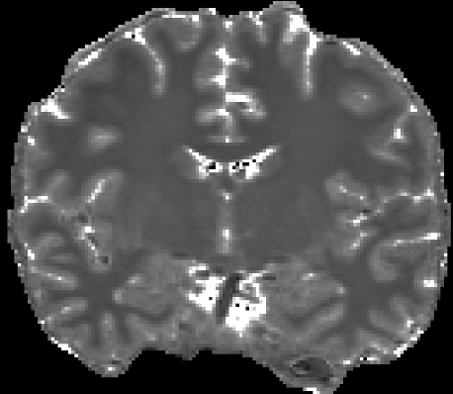}
			\\
			(a) & (b) & (c) & (d) & (e) \\
	\end{tabular}
		}
		\caption{PSACNN training images for a training subject $i$: (a) Training MPRAGE ($A_{(i)}$), (b) Training Label Image~($Y_{(i)}$), (c) $\rho_{(i)}$~(PD) map, (d) ${T_1}_{(i)}$, (e) ${T_2}_{(i)}$.}
		\label{fig:atlas}
	\end{figure}
Let $\mathcal{A} = \{A_{(1)}, A_{(2)}, \ldots, A_{(M)}\}$ be a collection of
$M$ $T_1$-w images with a corresponding expert manually labeled image set
$\mathcal{Y} = \{Y_{(1)}, Y_{(2)}, \ldots Y_{(M)}\}$. The paired collection
$\{\mathcal{A}, \mathcal{Y}\}$ is referred to as the
training image set. Let $\mathcal{A}$ be acquired
with a pulse sequence $\Gamma_A$, where $\Gamma_{A}$ is typically
MPRAGE that presents a good gray-white matter contrast. For PSACNN,
we require that in addition to $\{\mathcal{A}, \mathcal{Y}\}$, we have $\mathbf{\mathcal{B}}=[\boldBeta_{(1)}, \boldBeta_{(2)},
\ldots, \boldBeta_{(M)}]$; the corresponding NMR
parameter maps for each of the $M$ training subjects.
For each $i\in \{1, \ldots, M\}$ we have $\boldBeta_{(i)} = [{\rho}_{(i)},
{T_1}_{(i)}, {T_2}_{(i)}]$, where ${\rho}_{(i)}$ is a map of proton
densities, and ${T_1}_{(i)}$ and ${T_2}_{(i)}$ store the longitudinal~($T_1$),
and transverse~($T_2$) relaxation time maps respectively. Most training sets
do not acquire or generate the NMR maps. In case they are not available,
we outline an image synthesis procedure in Section~\ref{sec:betamap} to
generate these from available $T_1$-w MPRAGE images $A_{(i)}$.
Example images from $\{\mathcal{A}, \mathcal{Y}, \mathcal{B}\}$ are shown in
Fig.~\ref{fig:atlas}.

\begin{figure}[!ht]
	\begin{center}
 {\includegraphics[width=1\textwidth]{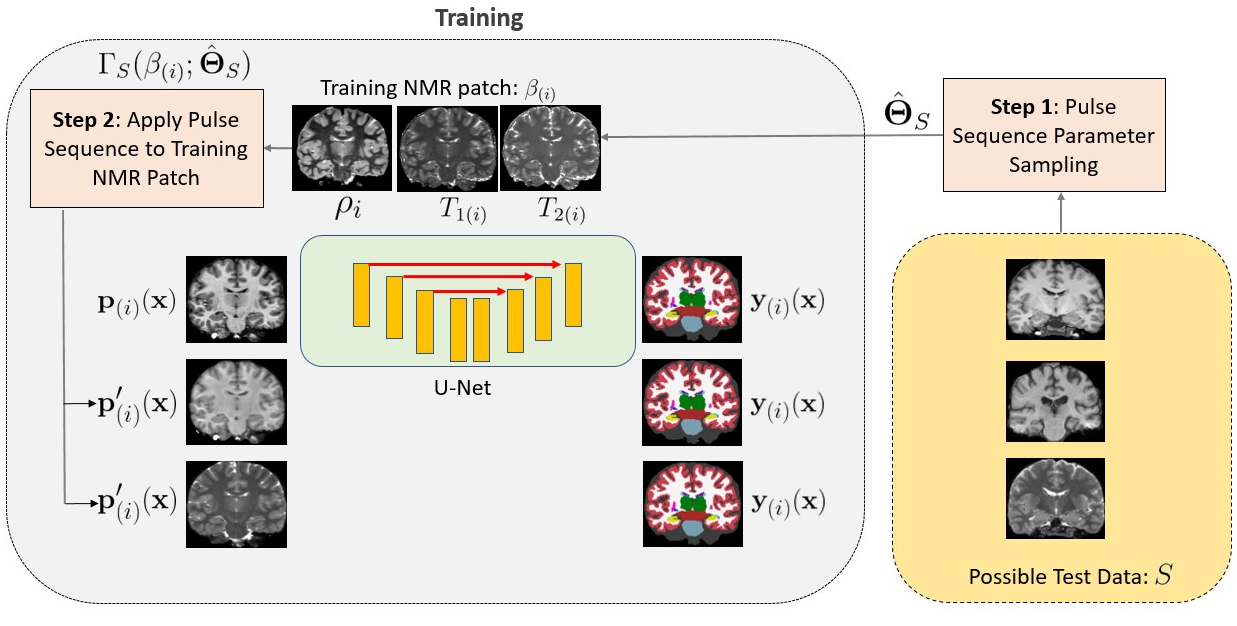}}
	\end{center}
	\caption{Workflow of PSACNN training.}
	\label{fig:method}
\end{figure}

 We extract patches $\boldp_{(i)}(\boldx)$ of size $96\times96\times96$ from voxels $\boldx$ of $A_{(i)}$. $\boldp_{(i)}(\boldx)$ is paired with a corresponding patch $\boldy_{(i)}(\boldx)$ extracted from the label image $Y_{(i)}$. Training
pairs of $[\boldp_{(i)}(\boldx); \boldy_{(i)}(\boldx)]$ constitute the unaugmented data used in training. A CNN trained on only available
unaugmented patches will learn to segment images with the exact same contrast as the training MPRAGE images and will not generalize to test images of a different contrast. We use the approximate forward models of pulse sequences
as described in Section~\ref{sec:estimation} to generate synthetic images that appear to have been imaged by that pulse sequence.

Consider a pulse sequence $\Gamma_S$, which could be MPRAGE, FLASH, SPGR, or T2-SPACE, for each of which we have established a three-parameter space.
We uniformly sample the $[\theta_0, \theta_1, \theta_2]$ grid that is created for $\Gamma_S$ as described in Section~\ref{sec:estimation} to get a candidate
set $\hat{\boldTheta}_{S}$. Next, we extract $96\times96\times96$-sized NMR parameter patches from $\boldBeta_{(i)} \in \mathcal{B}$ where the NMR patch at voxel $\boldx$ is denoted by $[\rho_{(i)}(\boldx),  {T_1}_{(i)}(\boldx),{T_2}_{(i)}(\boldx)]$.
We create an augmented patch $\boldp'_{(i)}(\boldx)$ by applying the test pulse sequence equation $\Gamma_S$, with the sampled pulse sequence parameter set $\hat{\boldTheta}_{S}$, to training NMR patches as shown in Eqn.~\ref{eq:synthpatch}:
\begin{equation}
\boldp'_{(i)}(\boldx) = \Gamma_S([\rho_{(i)}(\boldx), {T_{1}}_{(i)}(\boldx),{T_{2}}_{(i)}(\boldx)]; \hat{\boldTheta}_{S}).
\label{eq:synthpatch}
\end{equation}
The synthetic patch therefore has the anatomy from the $i^{\textrm{th}}$ training subject, but the image contrast of a potential test pulse sequence $\Gamma_S$ with $\hat{\boldTheta}_{S}$ as its parameters, as shown in Fig.~\ref{fig:method}.
The synthetic patch $\boldp'_{(i)}(\boldx)$ is paired with the corresponding label patch $\boldy_{(i)}(\boldx)$ extracted from $Y_i$ to create the augmented pair that is introduced in the training. The CNN, therefore, learns weights such that it maps both
$\boldp'_{(i)}(\boldx)$ and $\boldp_{(i)}(\boldx)$ to the same label patch
$\boldy_{(i)}(\boldx)$, in effect learning to be invariant to differences in
image contrast. CNN weights are typically optimized by stochastic optimization
algorithms that use mini-batches composed of training samples to update the
gradients of the weights in a single iteration. In our training, we construct a
mini-batch comprising four training samples: one sample of the original,
unaugmented MPRAGE patch, and one each of synthetic FLASH/SPGR, synthetic
T2-SPACE, and a synthetic MPRAGE patch. Each synthetic patch is generated with
a parameter set~($\hat{\boldTheta}_{S}$) chosen from the 3D parameter space
grid of its pulse sequence~($\Gamma_S$). Over a single epoch, we cover all
parameter sets from the 3D parameter space grid of $\Gamma_S$ to train the CNN
with synthetic patches that are mapped to the same label patch forcing it
to learn invariance across different $\hat{\boldTheta}_{S}$.
Invariance across pulse sequences is achieved by including patches from all
four pulse sequences in each mini-batch. The training workflow is presented in
Fig.~\ref{fig:method}. On the right side of Fig.~\ref{fig:method}
are possible test images that were acquired by different pulse sequences. On
the left side of Fig.~\ref{fig:method} are the synthetic training images
generated by applying the approximate pulse sequence equations to training NMR
maps shown on the top center. All synthetic images are matched with the same
label map at the output of the network. The patches are shown as 2D slices for
the purpose of illustration.

\subsection{PSACNN Network Architecture}
\label{sec:network}
We use the U-Net~\citep{unet2015} as our preferred voxel-wise segmentation
CNN architecture, but any sufficiently powerful fully convolutional architecture
can be used. Graphical processing unit~(GPU) memory constraints
prevent us from using entire $256\times256\times256$-sized images as input
to the U-Net and so we use $96\times96\times96$-sized patches as inputs.
Our U-Net instance has five pooling layers and corresponding five upsampling layers. These parameters were constrained by the largest model size we could fit in the GPU memory. Prior to each pooling layer are two convolutional
layers with filter size of $3\times3\times3$, ReLu activation, and batch
normalization after each of them. The number of filters in the first convolutional layers is 32, and it doubles after each pooling layer.
The last layer has a softmax activation with  $L=41$ outputs. The U-Net is trained to minimize a soft Dice-based loss averaged over the whole batch. The loss is shown in Eqn.~\ref{eq:loss}, where $\boldx$ denotes the voxels present in each label patch $\boldy_{true}(b)$ and $\boldy_{pred}(b)$ of the $b^\textrm{th}$ training sample in a batch of $N_B$ samples. During training $\boldy_{pred}(b)$ is a $96\times96\times96\times L$-sized tensor with each voxel recording the softmax probability of it belonging to a particular label. $\boldy_{true}(b)$
is a similar-sized one-hot encoding of the label present at each of the voxels.
The 4D tensors are flattened prior to taking element-wise dot products and
norms.
\begin{equation}
\textrm{Loss}(\boldy_{true}, \boldy_{pred}) = \frac{1}{N_B}\sum_{b}(1 - \frac{2\sum _{\boldx}\boldy_{true}(b)\cdot\boldy_{pred}(b)}{\sum_{\boldx}(||\boldy_{true}(b)||^2 + ||\boldy_{pred}(b)||^2 )})
\label{eq:loss}
\end{equation}
We use the Adam optimization algorithm to minimize the loss. Each epoch
uses $600$K augmented and unaugmented training pairs extracted from a
training set of 16 subjects out of a dataset with 39 subjects that were
acquired with 1.5~T Siemens Vision MPRAGE acquisitions~(TR=9.7~ms, TE=4~ms TI=20~ms, voxel-size=$1\times1\times1.5$~mm$^3$) with expert manual labels done as per the protocol described by~\cite{buckner2004protocol}. We refer to this dataset as the \textit{Buckner} dataset in the rest of the manuscript.
After each training epoch, the trained model is a applied to a validation
dataset that is generated from 3 of the remaining 23~(= 39 - 16) subjects.
Training is stopped when the validation loss does not decrease any more, which is usually around 10 epochs. For prediction, we sample overlapping $96\times96\times96$-sized patches from the test image with a
stride of $32$. We apply the trained network to these patches. The resultant soft predictions~(probabilities) are averaged in the overlapping regions and the class with the maximum probability is chosen as the hard label at each voxel.

\subsection{Synthesis of Training NMR Maps $\mathbf{\mathcal{B}}$}
\label{sec:betamap}
As described in Section~\ref{sec:training}, PSACNN requires the training dataset to be the collection $\{\mathcal{A}, \mathcal{Y}, \mathcal{B}\}$. But most training datasets only contain $\{\mathcal{A}, \mathcal{Y}\}$ and do not acquire or estimate the NMR parameter maps
$\mathcal{B}$. In this section, we will describe a procedure to synthesize
$\mathcal{B}$ maps from the available image set $\mathcal{A}$.

We use a separately acquired dataset of eight subjects, denoted
as dataset $\mathcal{F}$, with multi-echo FLASH~(MEF) images with four flip angles $3^\circ$, $5^\circ$, $10^\circ$, and $20^\circ$. Proton density ($\rho$) and $T_1$ maps were calculated by fitting the known imaging equation for FLASH to the acquired four images~\citep{Fischl2004}. The $\rho$ values are estimated up to a constant scale. Note that this dataset is completely distinct from the PSACNN training dataset $\mathcal{\{A, Y\}}$ with no overlapping subjects.
For each subject $j \in \{1,\ldots 8\}$ in $\mathcal{F}$, we first estimate
the three mean tissue class intensities $S_c$, $S_g$, and $S_w$.
We then estimate the MEF parameters~($\hat{\boldTheta}_{{MEF}(j)} = [\theta_0, \theta_1, \theta_2]$) using our approximation of the FLASH equation in Eqn.~\ref{eq:flash_approx2} and solving the equation system in Eqns.~\ref{eq:sc}--\ref{eq:sw}.
Next, the voxel intensity ${S_{MEF}}_{(j)}(\boldx)$ for each voxel $\boldx$, using the approximate FLASH imaging equation
in Eqn.~\ref{eq:flash_approx2} and the estimated $\hat{\boldTheta}_{{MEF}(j)} = [\theta_0, \theta_1, \theta_2]$
gives us Eqn.~\ref{eq:flash_map}:
\begin{equation}
 \log({S_{MEF}}_{(j)}(\boldx)) \approx \theta_0 + \log(\rho_{(j)}(\boldx)) + \frac{\theta_1}{{T_1}_{(j)}(\boldx)} + \frac{\theta_2}{{T_2}_{(j)}(\boldx)}.
\label{eq:flash_map}
\end{equation}
The only unknown in Eqn.~\ref{eq:flash_map} is ${T_2}_{(j)}(\boldx)$, which
we can solve for. This forms $\mathcal{B}_F = \{[\rho_{(j)}, {T_1}_{(j)}, {T_2}_{(j)}]\}$ for each subject in $\mathcal{F}$.

Next, we consider each of the PSACNN training images $A_{(i)}$ in  $\mathcal{A}$. We use the approximate forward model of the $\Gamma_A$
sequence~(typically MPRAGE) and the pulse sequence parameter estimation procedure in Section~\ref{sec:estimation} to estimate $\Gamma_A$ imaging parameters $\hat{\boldTheta}_{{A}(i)}$ for all PSACNN training images $A_{(i)}$. We use the estimated $\hat{\boldTheta}_{{A}(i)}$ as parameters in the $\Gamma_A$
equation and apply it to the NMR maps $B_F$ of dataset $\mathcal{F}$ as shown in Eqn.~\ref{eq:synmprage}, to generate synthetic $\Gamma_A$ images of subjects in dataset $\mathcal{F}$.
\begin{equation}
{S_{synA}}_{(j,i)} = \Gamma_{A}([\rho_{(j)}, {T_1}_{(j)}, {T_2}_{(j)}];\hat{\boldTheta}_{{A}(i)}).
\label{eq:synmprage}
\end{equation}

For each PSACNN training  image $A_{(i)}$, we have eight synthetic ${S_{synA}}_{(j,i)}$, $j \in \{1, \ldots, 8\}$ of subjects in dataset $\mathcal{F}$ with the same imaging parameter set $\hat{\boldTheta}_{{A}(i)}$ and therefore, the same image contrast as $A_{(i)}$. We pair up the synthetic images and proton density maps $\{{S_{synA}}_{(j,i)}, \rho_{(j)}\}$ to create a training dataset to estimate proton density map for subject $i$ in the PSACNN training set. We extract $96\times96\times96$-sized patches from the
synthetic images and map them to patches from the proton density maps.
A U-Net with specifications similar to that described in Section~\ref{sec:training} is used to predict proton density patches from
synthetic image patches, the only difference being the final layer that outputs $96\times96\times96\times1$-sized continuous valued output while
minimizing the mean squared error loss. The trained U-Net is applied to patches extracted from $A_{(i)}$ to synthesize $\rho_{(i)}$. The image contrast of the synthetic $\Gamma_A$ images ${S_{synA}}_{(j,i)}$ from dataset $\mathcal{F}$ is the same as $A_{(i)}$ by design, which makes the training and test intensity properties similar and the trained U-Net can be applied correctly to patches from $A_{(i)}$.
Two other similar image synthesis U-Nets are trained for synthesizing ${T_{1}}_{(i)}$ and ${T_{2}}_{(i)}$ from $A_{(i)}$. Thus, to estimate $[\rho_{(i)},{T_{1}}_{(i)}, {T_{2}}_{(i)}]$ for subject $i$ of the PSACNN
training dataset, we need to train three separate U-Nets. In total, for $M$
PSACNN training subjects, we need to train $3M$ U-Nets, which is computationally
expensive, but a one-time operation to permanently generate the $\mathcal{B}$ NMR maps that are necessary for PSACNN, from available $\mathcal{A}$ images.
It is possible to train only 3 U-Nets for all $M$ images from $\mathcal{A}$
as they have the exact same acquisition, however their estimated parameters are
not exactly equal and have a small variance. Therefore, we chose to train 3 U-Nets per image to ensure subject-specific training. Our expectation is that
generated NMR maps will be more accurate than those generated using a single
U-Net per NMR parameter map for all subjects.

The training dataset~(\textit{Buckner}) cannot be shared due to Institutional Review Board requirements. The trained PSACNN network and the prediction script
are freely available on the FreeSurfer development version repository on Github
(\url{https://github.com/freesurfer/freesurfer}).

\subsection{Application of PSACNN for 3~T MRI}
\label{sec:3tmri}
Our training dataset of $\{\mathcal{A}, \mathcal{Y}, \mathcal{B}\}$ was obtained on a 1.5~T Siemens Vision scanner. The tissue NMR parameters, $\rho$, $T_1$, $T_2$ are different at 3~T field strength than 1.5~T.
This might lead to the conclusion that it is not possible to create synthetic 3~T images during PSACNN augmentation as, in principle, we cannot apply 3~T
pulse sequence imaging equations to 1.5~T NMR maps. Ideally, we would prefer
to have a training dataset with $\mathcal{B}$ maps at both 1.5~T and 3~T, which unfortunately is not available to us. However, our imaging equation approximations allow us to frame a 3~T imaging equation as a 1.5~T imaging equation with certain assumptions about the relationship between
1.5~T and 3~T NMR parameters.
For instance, consider the FLASH approximate imaging equation in Eqn.~\ref{eq:flash_approx2}. When estimating the imaging parameters
of this equation, we construct an equation system~(for 1.5~T in Eqn.~\ref{eq:1p5_1}, reframed as Eqn.~\ref{eq:1p5_2}), as described in Section~\ref{sec:estimation}.
We assume that the 3~T FLASH sequence will have the same equation as the 1.5~T FLASH. This is a reasonable assumption because the sequence itself need not change with the field strength.

\begin{align}
\begin{bmatrix}
\log(s_c) \\
\log(s_g) \\
\log(s_w)
\end{bmatrix}
&=
\begin{bmatrix}
1 &  1/{T_1}_{c1.5} & 1/{T_2}_{c1.5} \\
1 &  1/{T_1}_{g1.5} & 1/{T_2}_{g1.5} \\
1 &  1/{T_1}_{w1.5} & 1/{T_2}_{w1.5}
\end{bmatrix}
\begin{bmatrix}
\theta_0 \\
\theta_1 \\
\theta_2
\end{bmatrix}
+
\begin{bmatrix}
\log(\rho_{c}) \\
\log(\rho_{g}) \\
\log(\rho_{w})
\end{bmatrix}
\label{eq:1p5_1}\\
\bolds & = \boldB_{1.5}\boldtheta_{1.5} + \boldrho \\
\bolds - \boldrho &  = \boldB_{1.5}\boldtheta_{1.5}
\label{eq:1p5_2}
\end{align}

We can express the equation system for a 3~T pulse sequence in Eqn.~\ref{eq:3t_1}, where $\boldB_{3}$ is $3\times3$-sized matrix similar to
$\boldB_{1.5}$ in Eqn.~\ref{eq:1p5_2}, but with mean tissue NMR values at 3~T, which can be obtained from previous studies~\citep{Wansapura1999}.
It is possible to estimate a $3\times3$-sized matrix $K$, where $\boldB_{3} = \boldB_{1.5}K$. $K$ is given by, $K = \boldB_{1.5}^{-1}\boldB_{3}$.
Now, given a 3~T image, we first calculate the mean tissue intensities $\bolds$.
The $\boldrho$ values are mean tissue proton density values and do not depend
on field strength. Therefore, we can write Eqn.~\ref{eq:3t_1}, that is similar
to Eqn.~\ref{eq:1p5_2}. Substituting $\boldB_{3}$ in Eqn.~\ref{eq:3t_1}, we get Eqn.~\ref{eq:3t_2} and reframe it as Eqn.~\ref{eq:3t_3}.
\begin{align}
\bolds - \boldrho & = \boldB_{3}\boldtheta_{3},
\label{eq:3t_1}\\
& = (\boldB_{1.5}K)\boldtheta_{3},
\label{eq:3t_2}\\
& = \boldB_{1.5}(K\boldtheta_{3}).
\label{eq:3t_3}
\end{align}
Therefore, we can treat the parameters for 3~T~($\boldtheta_{3}$) as linearly
transformed~(by $K^{-1}$) parameters $\boldtheta_{1.5}$. In essence, we can thus treat 3~T images as if they have been acquired with a 1.5~T scanner~(with 1.5~T NMR tissue values) but with a parameter set ${\boldtheta_{3}} = K^{-1}\boldtheta_{1.5}$.

However, while applying a 3~T equation to 1.5~T NMR maps to generate a
synthetic 3~T image, the above relationship of $\boldB_{3} = \boldB_{1.5}K$
implies a stronger assumption that $[1, 1/{T_1}_{v3}, 1/{T_2}_{v3}]^T = K^T[1,
1/{T_1}_{v1.5},1/{T_2}_{v1.5}]^T$, where the $[1, 1/{T_1}_{v3},1/{T_2}_{v3}]$ is a row vector with NMR values at any voxel $v$ at 3~T and $[1, 1/{T_1}_{v1.5},
1/{T_2}_{v1.5}]$ is a row vector of NMR values at 1.5~T for the same anatomy.
$K$ will have to be estimated using a least squares approach using all the 1.5~T and 3~T values at all voxels, which are not available to us.
This linear relationship between 1.5~T and 3~T NMR parameters is likely
not theoretically valid, but nevertheless works well in practice as can be seen
from empirical results presented in Section~\ref{sec:results} and
Figs.~\ref{fig:thetaspace}(a)--(c), where we can see the estimated parameters
from a variety of 1.5~T and 3~T MPRAGE scans. These are restricted in a small
enough range because the contrast changes for MPRAGE between 1.5~T and 3~T are
not drastically different and a linear relationship between the two NMR
vectors is likely a good enough approximation.

\section{Experiments and Results}
\label{sec:results}
We evaluate the performance of PSACNN and compare it with unaugmented CNN~(CNN),
FreeSurfer segmentation~(ASEG)~\citep{fischl2004aseg},
SAMSEG~\citep{puonti2016samseg}, and MALF~\citep{wang2013malf}.
We divide the experiments into two halves. In the first half, described
in Section~\ref{sec:accuracy}, we evaluate segmentation accuracy as measured by Dice overlap on datasets with manual labels.
In the second half, described in Section~\ref{sec:consistency}, we evaluate the segmentation consistency of all
five algorithms. Segmentation consistency is evaluated by calculating
the coefficient of variation of structure volumes for the same set of subjects
that have been scanned at multiple sites/scanners or sequences.

\subsection{Training Set and Pre-processing}
As described in Section~\ref{sec:network}, we randomly chose a subset comprising 16 out of 39 subjects representing the whole age range from the \textit{Buckner} dataset These subjects have 1.5~T Siemens Vision MPRAGE acquisitions and corresponding expert manual label images and were used as our training dataset for CNN, PSACNN, and as the atlas set for MALF.
ASEG segmentation in FreeSurfer and SAMSEG are model-based methods that use
a pre-built probabilistic atlas prior. The atlas priors for ASEG
and SAMSEG are built from 39 and 20 subjects respectively, from the  \textit{Buckner} dataset.

For MALF, we use cross-correlation as the registration metric and the parameter choices specific to brain MRI segmentation and joint label fusion as suggested by the authors in~\citep{wang2013malf}.

For PSACNN, we use the same single trained network that is trained from real and augmented MPRAGE, FLASH, SPGR, and T2-SPACE patches  of the 16-subject
training data for all the experiments. We do not re-train it for a new test dataset or a new experiment.

All input images from the training and test datasets are brought into the FreeSurfer conformed space~(resampled to isotropic $1\times1\times1$ mm$^3$)
skullstripped and intensity inhomogeneity-corrected~\citep{N3} by the
pre-processing stream of FreeSurfer~\citep{Fischl2004}. The intensities
are scaled such that they lie in the $[0,1]$ interval.

For the unaugmented CNN specifically, we further standardize the input image intensities such that their white matter mean~(as identified by fitting a Gaussian mixture model to the intensity histogram) is fixed to 0.8. The CNN results were significantly worse on multi-scanner data without this additional scaling. We also tested using piece-wise linear intensity standardization~\citep{nyul1999mrm} on input images to the unaugmented CNN but the results were better for white matter mean standardization.
There is no such scaling required for the other methods including PSACNN.

\subsection{Evaluation of Segmentation Accuracy}
\label{sec:accuracy}
We evaluate all algorithms in three settings, where the test datasets have varying degrees of acquisition differences with the
training \textit{Buckner} dataset. Specifically, where the test dataset is acquired by:
\begin{itemize}
\item Same scanner~(Siemens Vision), same sequence~(MPRAGE) as the training dataset
\item Different scanner~(Siemens SONATA), same sequence~(MPRAGE with different parameters) as the training set.
\item Different scanner~(GE Signa), different sequence~(SPGR) as the training dataset.

\end{itemize}

\subsubsection{Same Scanner, Same Sequence Input}
\label{sec:buckner}

\begin{figure}[!ht] \tabcolsep 1pt
	\centerline{
		\begin{tabular}{cccc}
			\includegraphics[width=.25\textwidth]{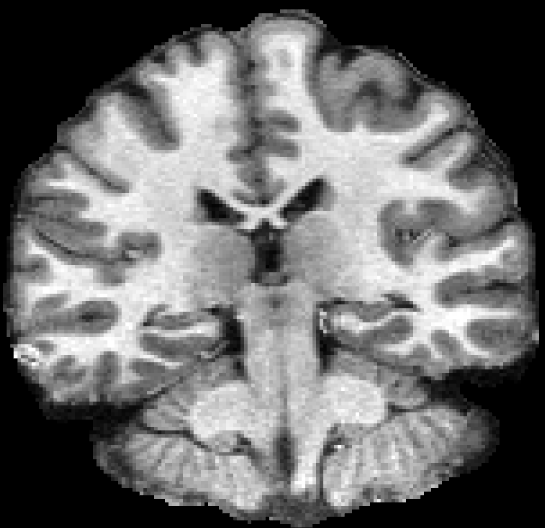} &
			\includegraphics[width=.25\textwidth]{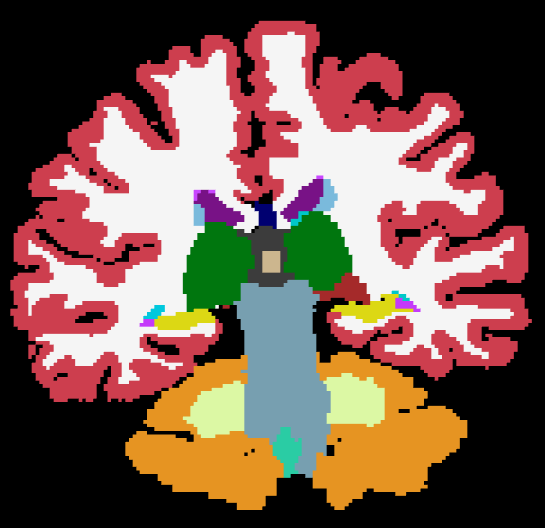} &
			\includegraphics[width=.25\textwidth]{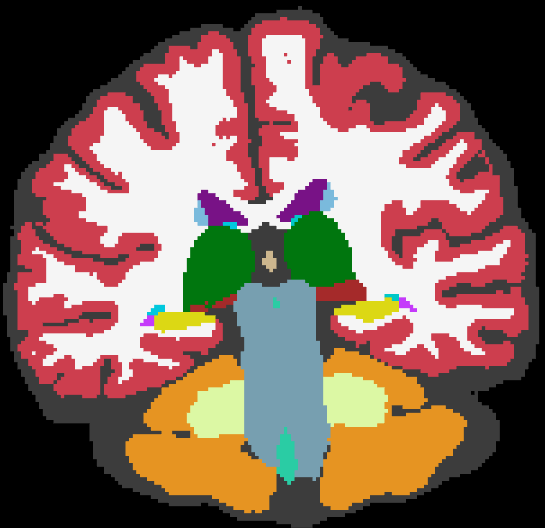} &
			\includegraphics[width=.25\textwidth]{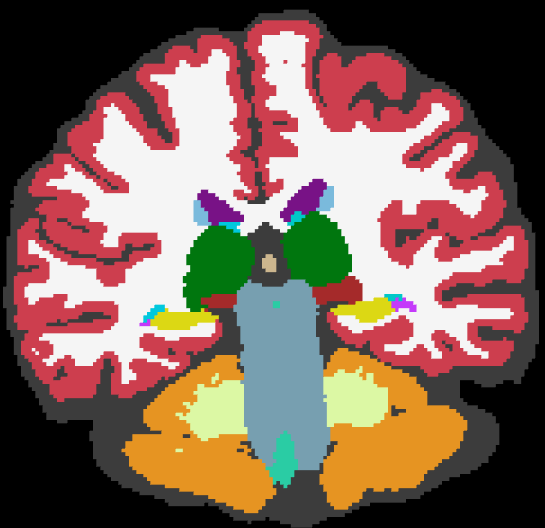}\\
			 (a) MPRAGE & (b) ASEG & (c) MALF & (d) PSACNN\\
			 \qquad&
			\includegraphics[width=.25\textwidth]{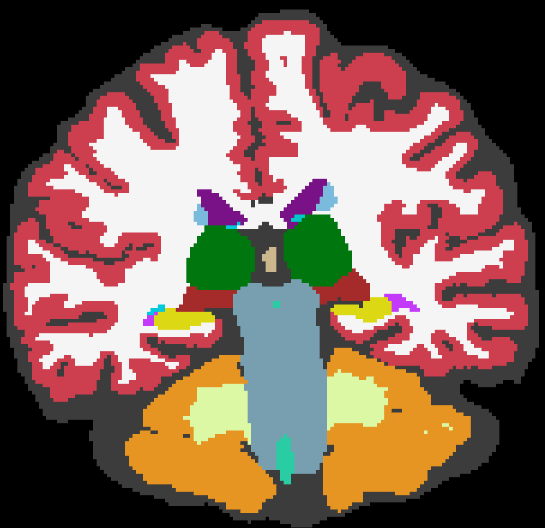}&
			\includegraphics[width=.25\textwidth]{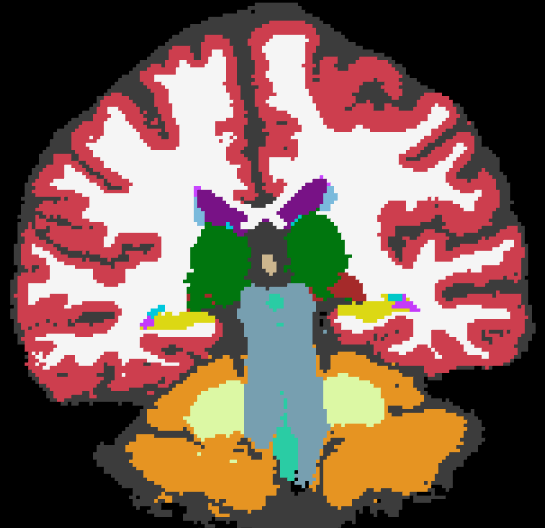}&
			\includegraphics[width=.25\textwidth]{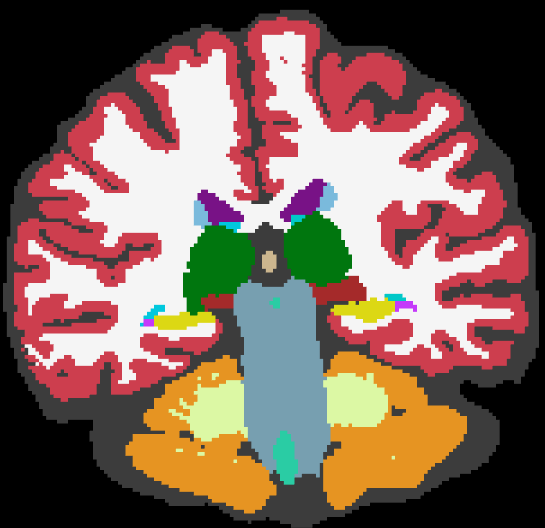}\\
			\qquad & (e) Manual & (f) SAMSEG & (g) CNN
	\end{tabular}
	}
	\caption{Input \textit{Buckner} MPRAGE with segmentation results from ASEG, SAMSEG, MALF, CNN, and PSACNN, along with manual segmentations.}
	\label{fig:bucknerseg}
\end{figure}

In this experiment we compare the performance of segmentation algorithms on
test data with the exact same acquisition parameters as the training data.
The test data consists of 20 subject MPRAGEs from the \textit{Buckner} data that
are independent of the 16 subjects used for training and 3 for validation
of the CNN methods.

\begin{figure}[ht!] \tabcolsep 1pt
	\centerline{
		\begin{tabular}{c}
			\includegraphics[width=\textwidth]{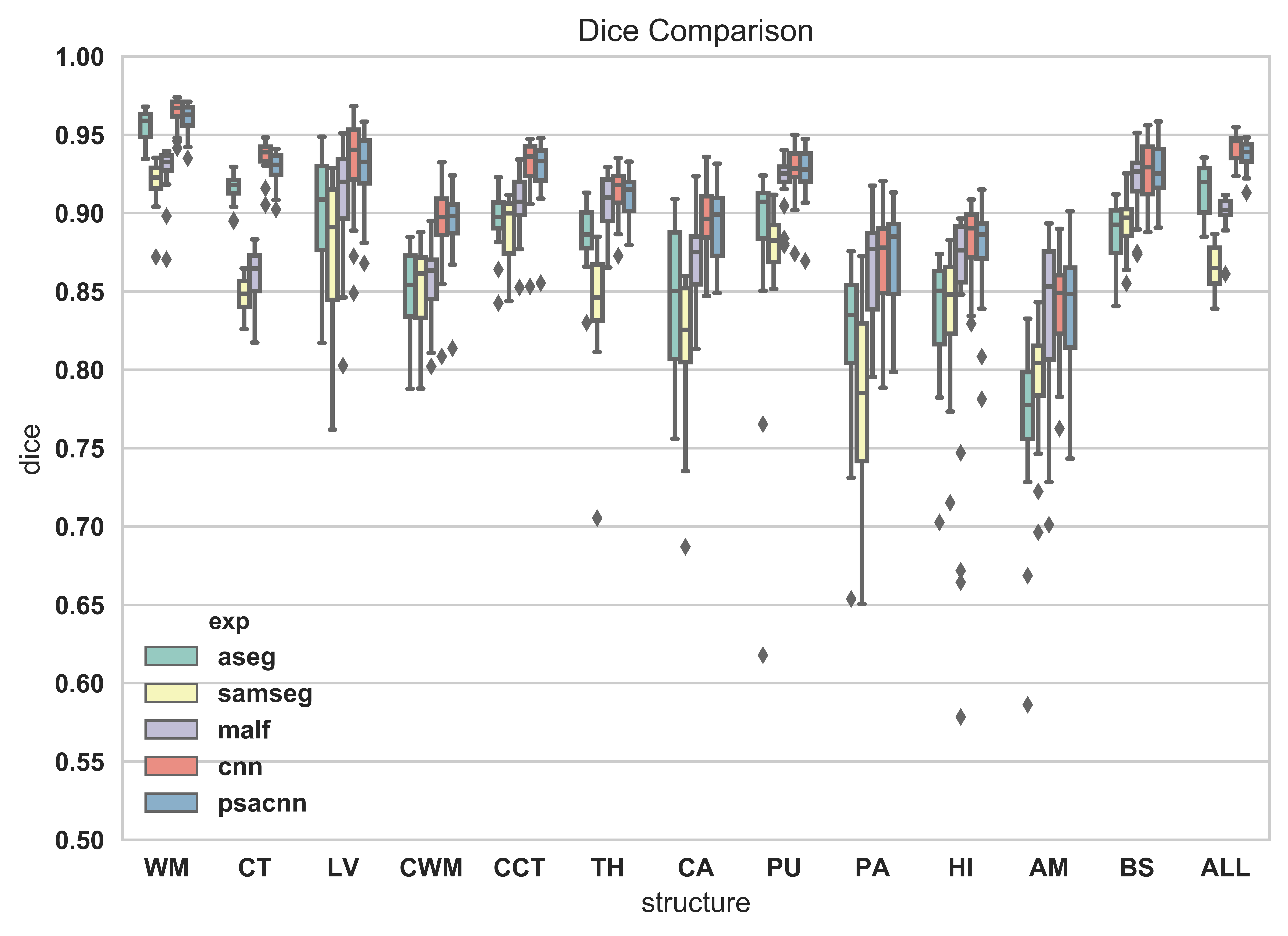}
		\end{tabular}
	}
	\caption{Dice coefficient boxplots of selected structures for all five methods on 20 subjects of the \textit{Buckner} dataset. Acronyms: white matter~(WM), cortex~(CT), lateral ventricle~(LV), cerebellar white matter~(CWM), cerebellar cortex~(CCT), thalamus~(TH), caudate~(CA), putamen~(PU), pallidum~(PA), hippocampus~(HI), amygdala~(AM), brain-stem~(BS), overlap for all structures~(ALL).}
\label{fig:bucknerdice}
\end{figure}

Figure~\ref{fig:bucknerseg} shows an example input test subject image, its
manual segmentation, and segmentations of the various algorithms. All
algorithms perform reasonably well on this dataset. We calculated the Dice coefficients of the labeled structures for all the algorithms~(see Fig.~\ref{fig:bucknerdice}).
From the Dice coefficient boxplots~(generated from 20 subjects) shown in Fig.~\ref{fig:bucknerdice} we observe that CNN~(red) and PSACNN~(blue) Dice overlap~(ALL Dice $=0.9396$) is comparable to each other and is significantly better~(paired t-test $p < 0. 05$) than ASEG~(green), SAMSEG~(yellow), and MALF~(purple) for many of the subcortical structures, and especially for whole white matter~(WM) and gray matter cortex~(CT) labels.
This suggests that large patch-based CNN methods do provide a more accurate
segmentation than state-of-the-art approaches like MALF and SAMSEG for
a majority of the labeled structures. But, since the training and test
acquisition parameters are exactly same, there is no significant difference between CNN and PSACNN.

\subsubsection{Different Scanner, Same Sequence Input}
\label{sec:siemens13}
\begin{figure}[!ht] \tabcolsep 1pt
	\centerline{
		\begin{tabular}{cccc}
			\includegraphics[width=.25\textwidth]{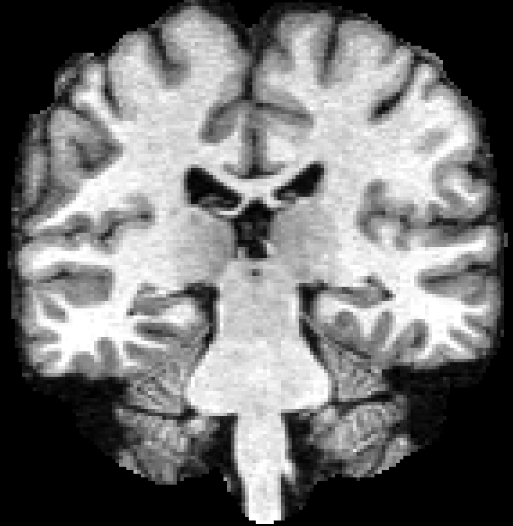} &
			\includegraphics[width=.25\textwidth]{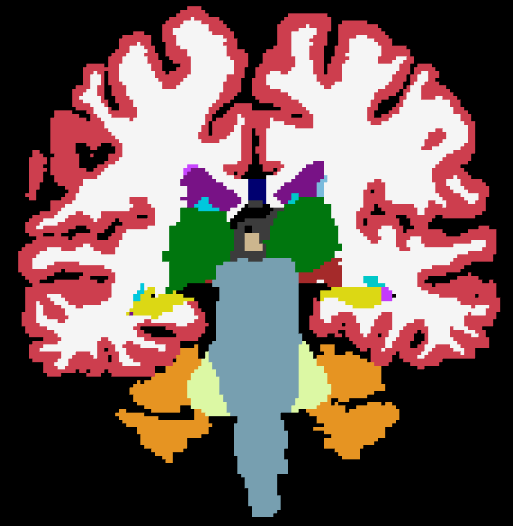} &
			\includegraphics[width=.25\textwidth]{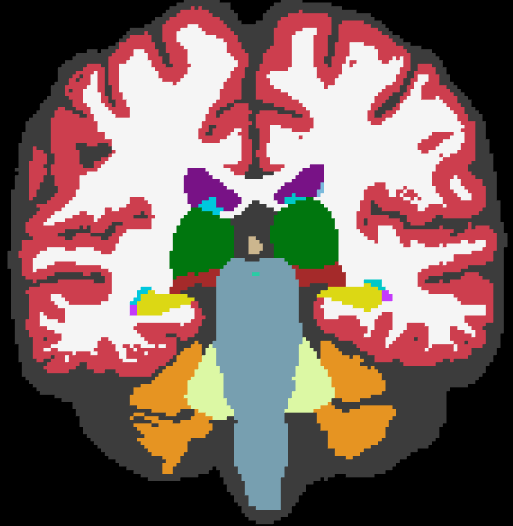} &
			\includegraphics[width=.25\textwidth]{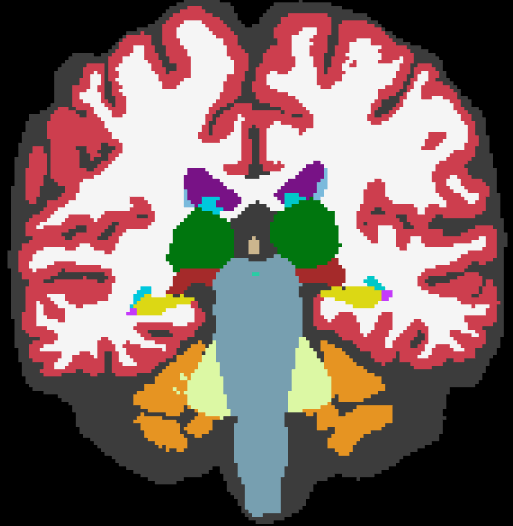}\\
			 (a) MPRAGE & (b) ASEG & (c) MALF & (d) PSACNN\\
			 \qquad&
			\includegraphics[width=.25\textwidth]{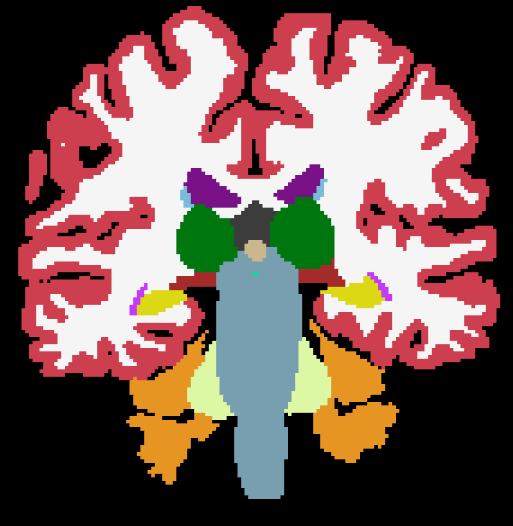}&
			\includegraphics[width=.25\textwidth]{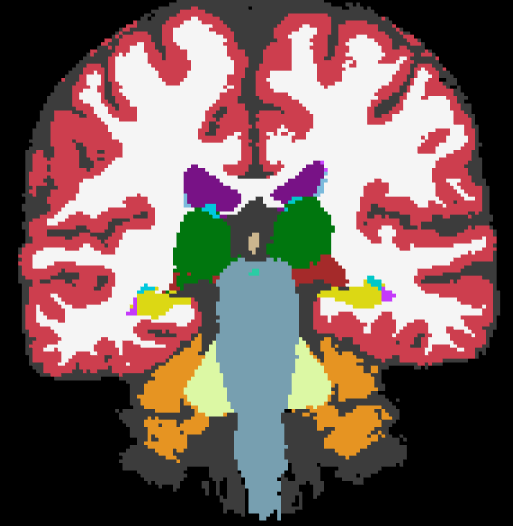}&
			\includegraphics[width=.25\textwidth]{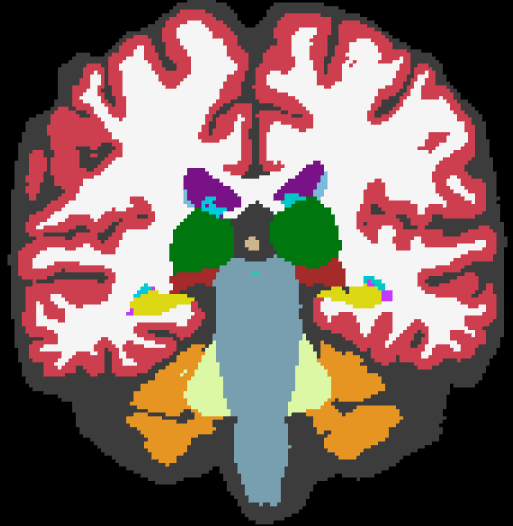}\\
			\qquad & (e) Manual & (f) SAMSEG & (g) CNN
	\end{tabular}
	}
	\caption{Input \textit{Siemens13} MPRAGE with segmentation results from ASEG, SAMSEG, MALF, CNN, and PSACNN, along with manual segmentations.}
	\label{fig:siemens14seg}
\end{figure}
\begin{figure}[t] \tabcolsep 1pt
	\centerline{
		\begin{tabular}{c}
			\includegraphics[width=\textwidth]{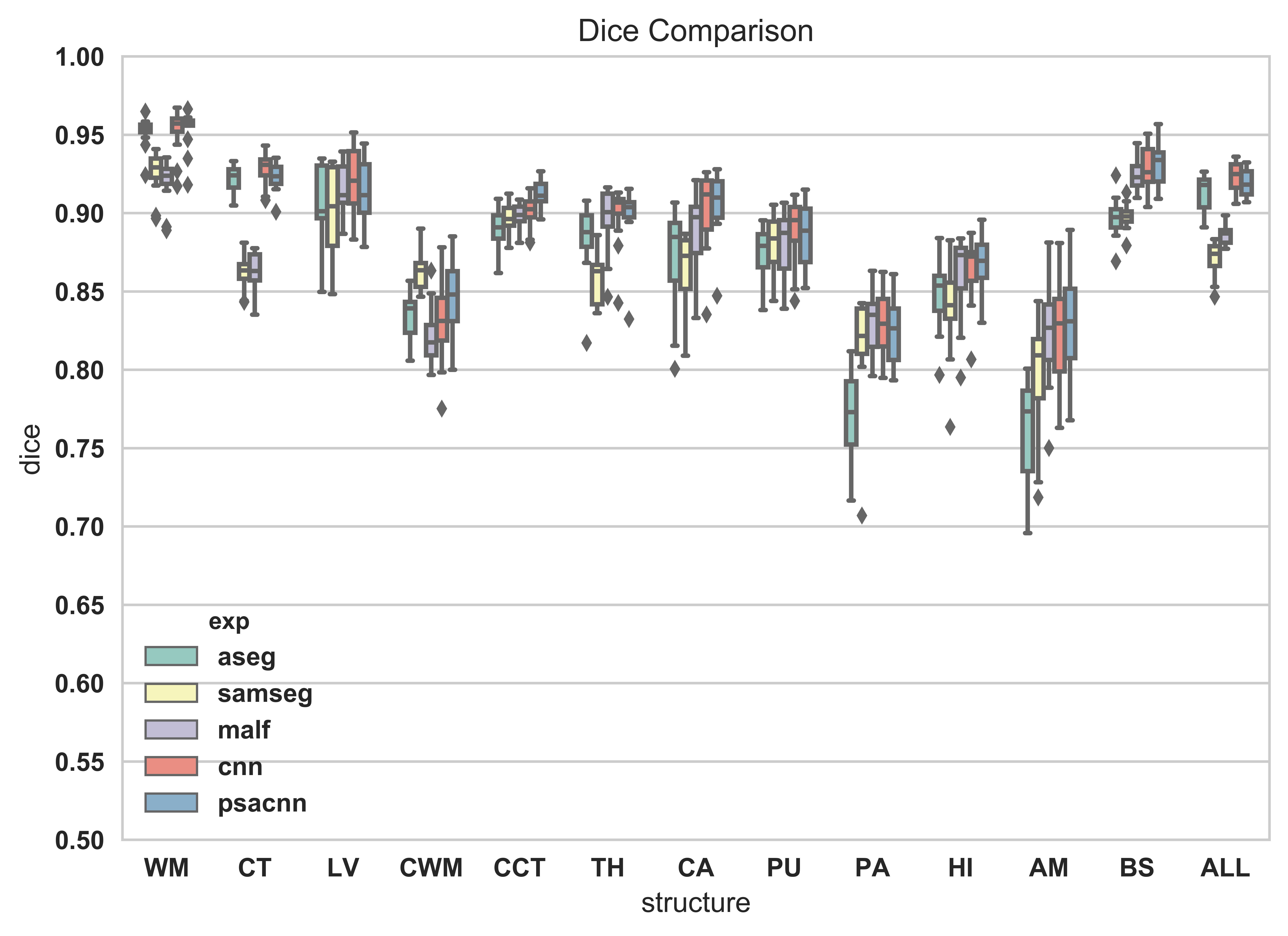}
		\end{tabular}
	}
	\caption{Dice evaluations on \textit{Siemens13} dataset. For acronyms refer to Fig.~\ref{fig:bucknerdice} caption.}
	\label{fig:siemens14dice}
\end{figure}

In this experiment we compare the accuracy of PSACNN against other methods on the \textit{Siemens13} dataset that comprises 13 subject MPRAGE scans acquired on a 1.5~T Siemens SONATA scanner with a similar pulse sequence as the training data. The \textit{Siemens13} dataset has expert manual segmentations generated with the same protocol as the training data~\citep{buckner2004protocol}.
Figure~\ref{fig:siemens14seg} shows the segmentation results for all the algorithms. The image contrast~(see Fig.~\ref{fig:siemens14seg}(a)) for the \textit{Siemens13} dataset is only slightly different from the training \textit{Buckner} dataset. Therefore, all algorithms
perform fairly well on this dataset. Figure~\ref{fig:siemens14dice} shows the Dice overlap boxplots calculated over 13 subjects. Similar to the previous experiment, the unaugmented CNN and PSACNN have a superior Dice overlap compared to other methods~(statistically significantly better for five of the nine labels shown), but are comparable to each other as the test acquisition is not very different from the training acquisition.

\subsubsection{Different Scanner, Different Sequence Input}
\label{sec:ge14}

\begin{figure}[!ht] \tabcolsep 0pt
	\centerline{
		\begin{tabular}{cccc}
			\includegraphics[width=.25\textwidth]{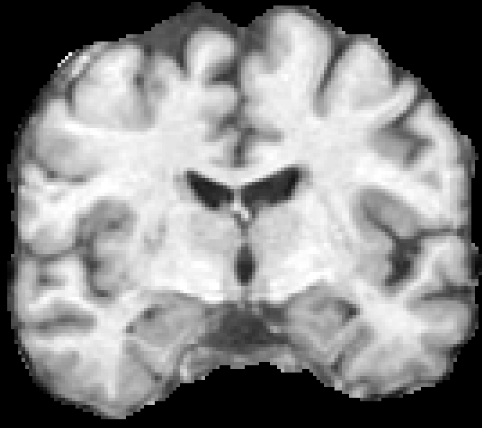} &
			\includegraphics[width=.25\textwidth]{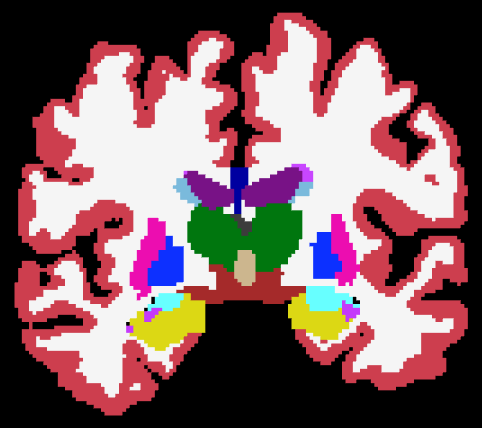} &
			\includegraphics[width=.25\textwidth]{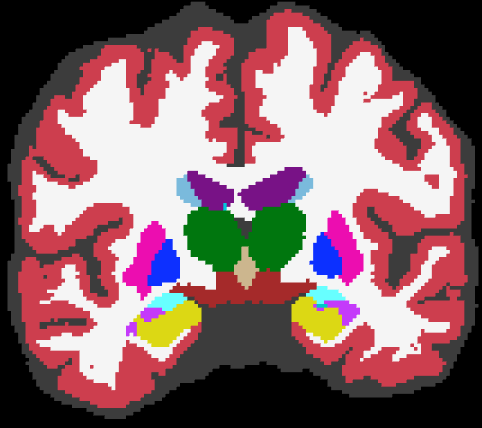} &
			\includegraphics[width=.25\textwidth]{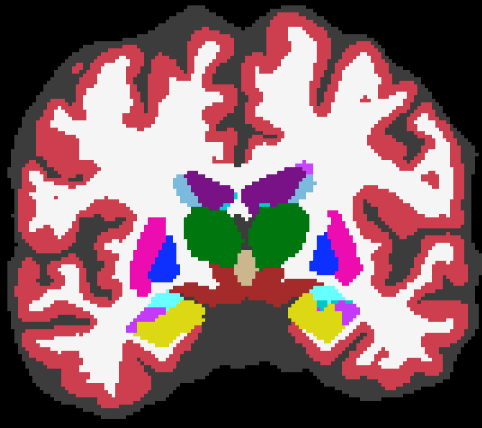}\\
			 (a) SPGR & (b) ASEG & (c) MALF & (d) PSACNN\\
			 \qquad&
			\includegraphics[width=.25\textwidth]{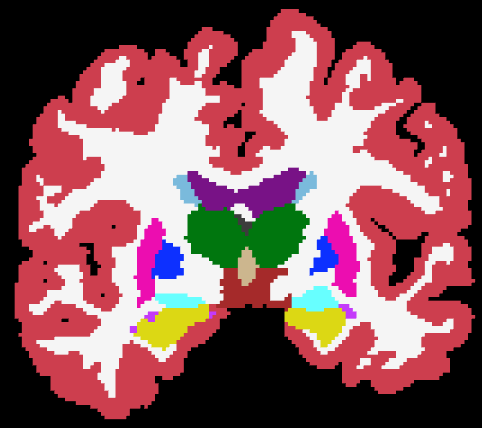}&
			\includegraphics[width=.25\textwidth]{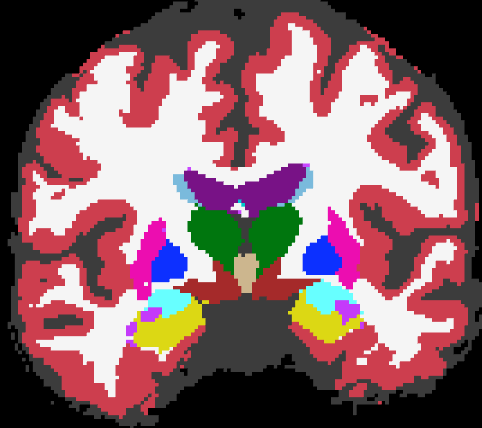}&
			\includegraphics[width=.25\textwidth]{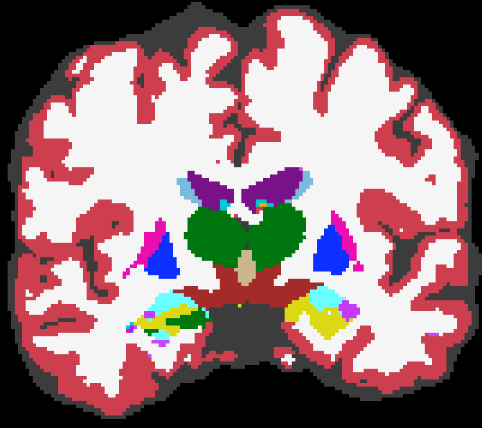}\\
			\qquad & (e) Manual & (f) SAMSEG & (g) CNN
	\end{tabular}
	}
	\caption{Input \textit{GE14} SPGR with segmentation results from ASEG, SAMSEG, MALF, CNN, and PSACNN, along with manual segmentations.}
	\label{fig:ge14seg}
\end{figure}

\begin{figure}[t] \tabcolsep 1pt
	\centerline{
		\begin{tabular}{c}
			\includegraphics[width=\textwidth]{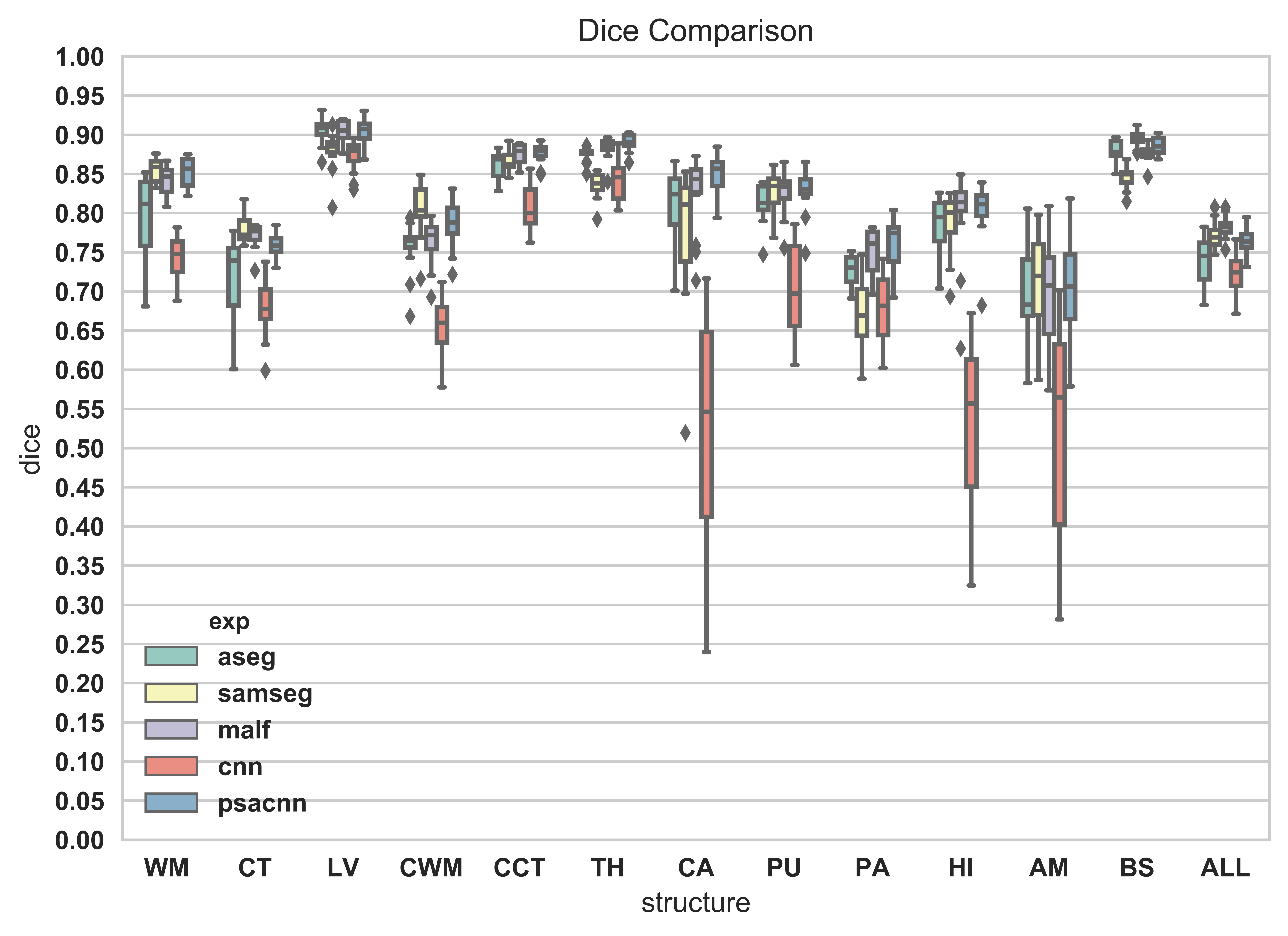}
		\end{tabular}
	}
	\caption{Dice evaluations on \textit{GE14} dataset. For acronyms refer to Fig.~\ref{fig:bucknerdice} caption.}
	\label{fig:ge14dice}
\end{figure}

In this experiment we compare the accuracy of PSACNN against other methods on
the $\textit{GE14}$ dataset comprising 14 subjects scanned on a 1.5~T GE Signa scanner with the SPGR~(spoiled gradient recalled) sequence~($TR=35~ms$, $TE=5$~ms, $\alpha=45^{\circ}$, voxel size=$0.9375\times0.9375\times1.5$~mm$^3$).
All the subjects have expert manual segmentations generated with the same protocol as the training data, but the manual labels are visibly
different than training data, with a much thicker cortex~(Fig.~\ref{fig:ge14seg}(e)). This is likely because the \textit{GE14} SPGR scans~(Fig.~\ref{fig:ge14seg}(a)) present a noticeably different GM-WM
image contrast than the MPRAGE training data. Consequently, as seen in Fig.~\ref{fig:ge14dice}, all methods show a reduced overlap, but the unaugmented CNN~(red boxplot) has the worst performance of all as it is unable to generalize to an SPGR sequence~(despite white matter peak intensity standardization).
Some obvious CNN errors are for small subcortical structures such as hippocampus~(yellow in Fig.~\ref{fig:ge14seg}(g)) labeled as thalamus~(green label). PSACNN~(ALL Dice=$0.7636$) on the other hand, is robust to the contrast change due to pulse sequence-based augmentation, and produces segmentations that are comparable to the state-of-the-art algorithms such as SAMSEG~(ALL Dice=$0.7708$) and MALF~(ALL Dice=$0.7804$) in accuracy, with 1--2 orders of magnitude lower processing time.

\subsection{Evaluation of Segmentation Consistency}
\label{sec:consistency}
The experiments described in Section~\ref{sec:accuracy} demonstrate that PSACNN
is a highly accurate segmentation algorithm and better or comparable to the
state-of-the-art segmentation algorithms. In this section, we evaluate the
segmentation consistency of all algorithms on three distinct
multi-scanner/multi-sequence datasets.

\subsubsection{Four Scanner Data}
\label{sec:fourscanner}
\begin{table}[!t]
	\caption{Mean (and std. dev.) of the coefficient of variation (in $\%$) of
	structure volumes	over all four datasets with all algorithms.}
	\label{tab:coeffvar_threescanners} \tabcolsep 0pt
	\vspace*{0.6em}
	\centerline{
	\resizebox{1\textwidth}{!} {
		\begin{tabular}{l c c c c c c c c c c c}
		  \toprule  & \hspace*{6.0ex} &
			\multicolumn{1}{c}{\textit{ASEG}} & \hspace*{4.0ex} &
			\multicolumn{1}{c}{\textit{SAMSEG}} & \hspace*{4.0ex} &
			\multicolumn{1}{c}{\textit{MALF}} & \hspace*{4.0ex} &
			\multicolumn{1}{c}{\textit{CNN}} & \hspace*{4.0ex} &
			\multicolumn{1}{c}{\textit{PSACNN}}
			\\
			 &&
			 \textbf{Mean} \textbf{~(Std)} &&
			 \textbf{Mean} \textbf{~(Std.)} &&
			 \textbf{Mean} \textbf{~(Std)} &&
			 \textbf{Mean} \textbf{~(Std.)} &&
			 \textbf{Mean} \textbf{~(Std.)} \\
			\cmidrule{3-3}
			\cmidrule{5-5} \cmidrule{7-7} \cmidrule{9-9} \cmidrule{11-11}
			 WM && 11.02 (7.19) && \0{}2.46 (2.14) && \0{}3.21 (1.23) && 17.27 (3.83) && \textbf{1.99} (1.19) \\
			 CT && 5.10 (3.75) && \0{}3.16 (1.30) && \0{}2.13 (0.73) && 7.21 (2.95) && \textbf{1.81} (1.30) \\
			 TH && 3.72 (1.49) && \0{}2.13 (0.62) && \0{}1.12 (0.55) && 6.08 (7.13) && \textbf{1.05} (0.62) \\
			 CA && 3.76 (2.10) && \textbf{\0{}1.04} (1.47) && \0{}2.42 (1.50) && 18.62 (11.05) && 2.15 (1.04) \\
			 PU && 7.42 (2.03) && \0{}2.99 (0.86) && \0{}2.03 (0.95) && 10.70 (4.16) && \textbf{1.05} (0.86) \\
			 PA && 5.22 (3.32) && \0{}8.52 (4.45) && \0{}3.83 (2.44) && 5.04 (3.37) && \textbf{3.11} (4.45) \\
			 HI && 3.18 (1.23) && \0{}2.45 (0.95) && \textbf{\0{}1.46} (0.88) && 16.28 (10.21) && 1.47 (0.54) \\
			 AM && 5.27 (2.20) && \0{}2.09 (1.09) && \textbf{\0{}1.99} (1.19) && 14.37 (6.75) && 2.16 (0.98) \\
			 LV && 5.61 (6.53) && \textbf{\0{}3.12} (5.90) && \0{}4.72 (6.26) && 10.40 (6.98) && 4.58 (6.72) \\
			\bottomrule
		 \end{tabular}
		  }
			}
	\vspace{0.5em}
	\centerline{
		\scriptsize{ \begin{minipage}{0.93\textwidth} \textbf{Bold} results show
		the minimum coefficient of variation. \end{minipage} } }
\end{table}

In this experiment we tested the consistency of the segmentation produced by
the various methods on four datasets acquired  from 13 subjects;
\begin{itemize}
\item \textit{MEF}: 3D Multi-echo FLASH, scanned on Siemens Trio 3~T scanner, voxel-size $1\times1\times1.33$ mm$^3$, sagittal acquisition, $TR$=20~ms, $TE$=1.8+1.82n~ms, flip angle 30$^\circ$.
\item \textit{TRIO}: 3D MPRAGE, scanned on Siemens Trio 3~T scanner, voxel-size $1\times1\times1.33$ mm$^3$, sagittal acquisition, $TR$=2730~ms, $TE$=3.44~ms, $TI$=1000~ms, flip angle 7$^\circ$.
\item \textit{GE}: 3D MPRAGE, scanned on GE Signa 1.5~T scanner, voxel-size $1\times1\times1.33$ mm$^3$, sagittal acquisition, $TR$=2730~ms, $TE$=3.44~ms, $TI$=1000~ms, flip angle 7$^\circ$.
\item \textit{Siemens}: 3D MPRAGE, scanned on Siemens Sonata 1.5~T scanner, voxel-size $1\times1\times1.33$ mm$^3$, sagittal acquisition, $TR$=2730~ms, $TE$=3.44~ms, $TI$=1000~ms, flip angle 7$^\circ$.
\end{itemize}


For each structure $l$ in each subject, we calculated the standard deviation~($\sigma_l$) of the segmented label volume over
all datasets. We divided this by the mean of the structure
volume~($\mu_l$) for that subject, over all the datasets.
This gives us $\frac{\sigma_l}{\mu_l}$, the coefficient of variation for
structure $l$ for each subject. In Table~\ref{tab:coeffvar_threescanners}, we
calculated the mean coefficient of variation~(in $\%$) over the 13 subjects in the Four Scanner Dataset. The lower the coefficient of variation, the more consistent is the segmentation algorithm in predicting the structure volume across multi-scanner datasets. From Table~\ref{tab:coeffvar_threescanners}, we can observe that PSACNN has the lowest coefficient of variation in five of the
nine structures and is the second lowest in three others. In the structures
where PSACNN has the lowest coefficient of variation, it is statistically significantly
lower ($p < 0.05$, Wilcoxon signed rank test) than the third placed method, but not the second-placed method. SAMSEG has the lowest coefficient of variation
for CA~(caudate) and LV~(lateral ventricle), whereas MALF
is the lowest for HI~(hippocampus) and  AM~(amygdala).

\begin{sidewaysfigure}[!ht] \tabcolsep 0pt
	\centerline{
		\begin{tabular}{ccc}
			\includegraphics[width=.33\textwidth]{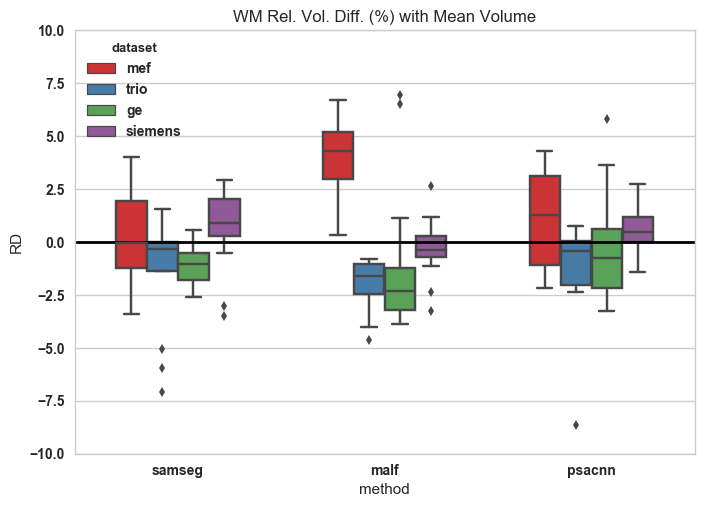} &
			\includegraphics[width=.33\textwidth]{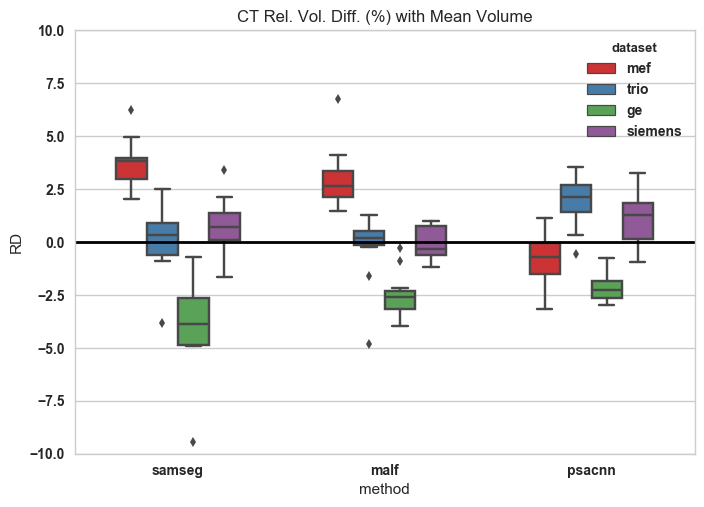} &
			\includegraphics[width=.33\textheight]{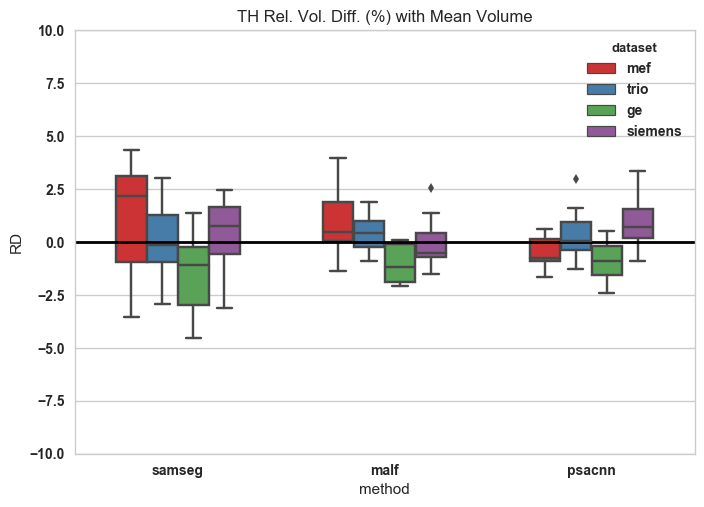}
			\\
			[-1.5em]
			{\textbf{(a)}}\hspace*{.25\textwidth} &
			{\textbf{(b)}}\hspace*{.25\textwidth} &
			{\textbf{(c)}}\hspace*{.25\textwidth}
			\\
	  	\includegraphics[width=.33\textwidth]{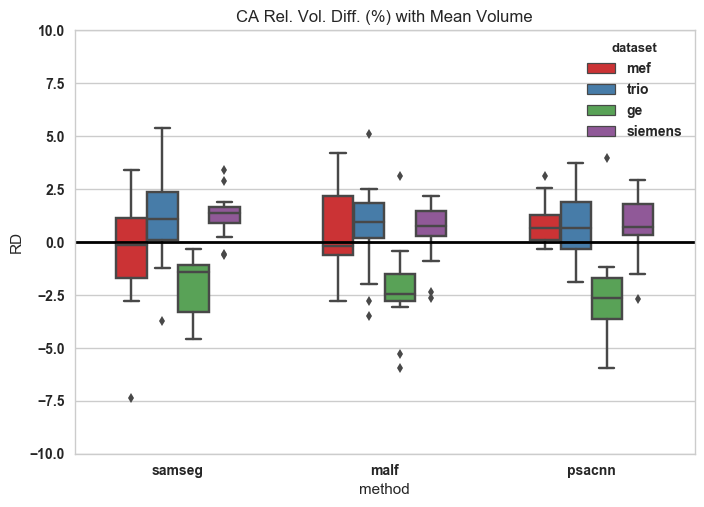} &
			\includegraphics[width=.33\textwidth]{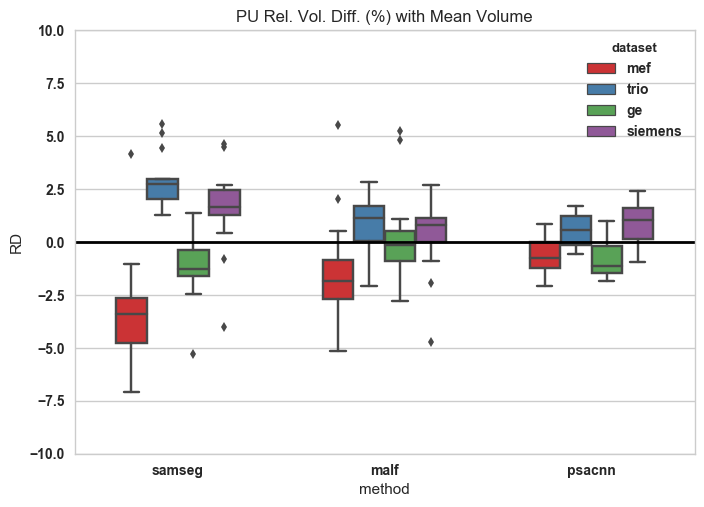} &
			\includegraphics[width=.33\textwidth]{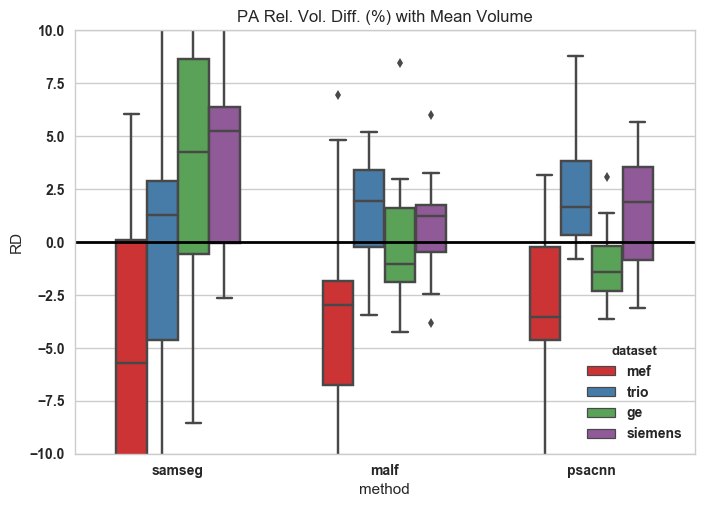} \\
			[-1.5em]
			{\textbf{(d)}}\hspace*{.25\textwidth} &
			{\textbf{(e)}}\hspace*{.25\textwidth} &
			{\textbf{(f)}}\hspace*{.25\textwidth}
			\\
			\includegraphics[width=.33\textwidth]{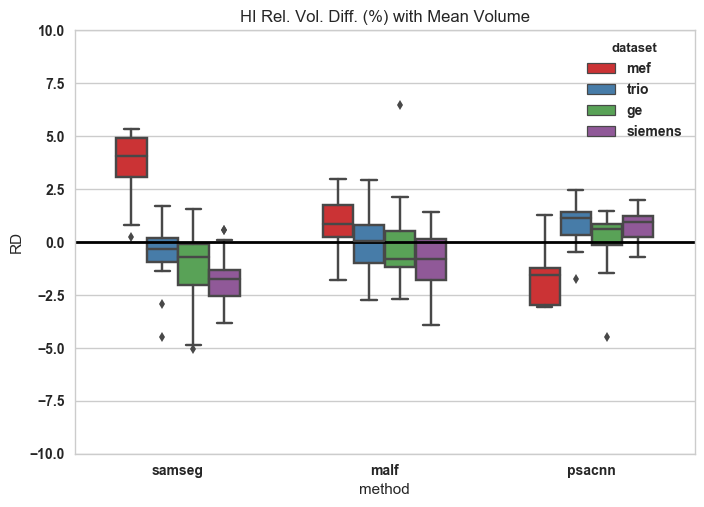} &
			\includegraphics[width=.33\textwidth]{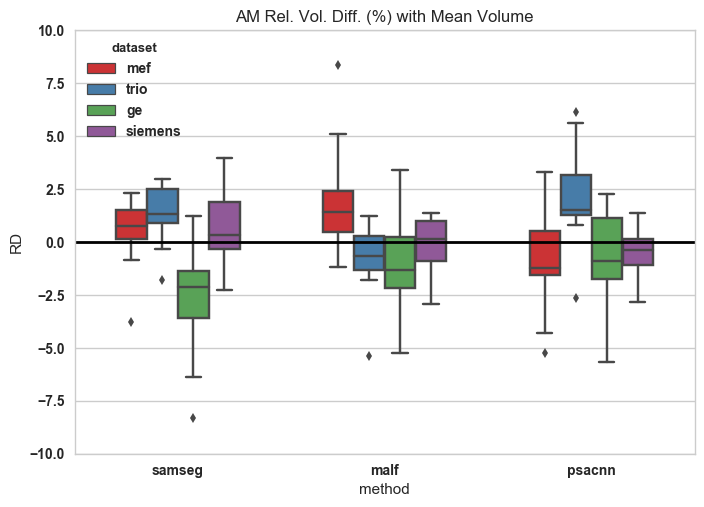} &
			\includegraphics[width=.33\textwidth]{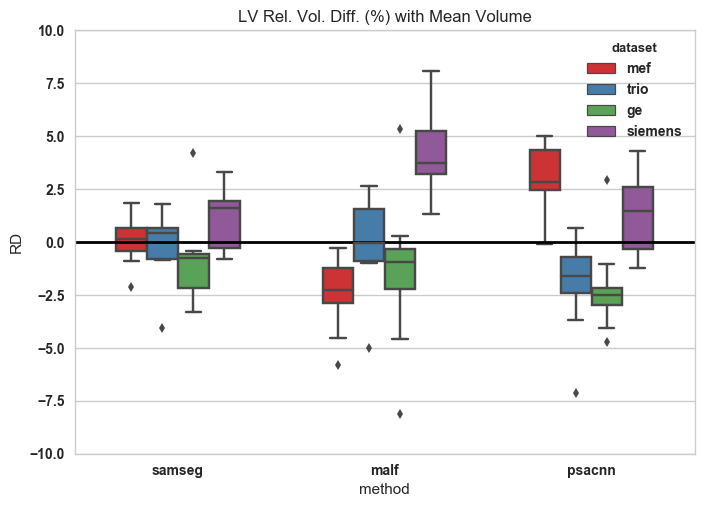}
			\\
			[-1.5em]
			{\textbf{(g)}}\hspace*{.25\textwidth} &
			{\textbf{(h)}}\hspace*{.25\textwidth} &
			{\textbf{(i)}}\hspace*{.25\textwidth}
	\end{tabular}
	}
	\caption{Signed relative difference of volumes of (a) WM, (b) CT, (c) TH, (d) CA, (e) PU, (f) PA, (g) HI, (h) AM, (i) LV from mean structure volume over all four datasets. Left group of boxplots shows SAMSEG, middle shows MALF, right-most shows PSACNN. The colors denote different datasets, \textit{MEF}~(red), \textit{TRIO}~(blue), \textit{GE}~(green), \textit{Siemens}~(purple)}
	\label{fig:rd_threescanners}
\end{sidewaysfigure}

Next, we calculated the signed relative difference of structure volume
estimated for a particular dataset with respect to the mean structure volume across all four datasets and show them as boxplots in Fig.~\ref{fig:rd_threescanners}~(\textit{MEF} in red, \textit{TRIO} in blue, \textit{GE} in green, \textit{Siemens} in purple.). Ideally, a segmentation algorithm would have zero bias and would produce the exact same segmentation for different acquisitions of the same subject. But the lower the bias, the more consistent is the algorithm across different acquisitions. Figures~\ref{fig:rd_threescanners}(a)--(i) show the relative difference
for each of the nine structures for SAMSEG, MALF, PSACNN. There is a small,
statistically significant bias (difference from zero, thick black line) for most
structures and algorithms, which suggests that there is a scope for further improvement for all algorithms. The relative differences also give us an
insight into which datasets show the most difference from the mean, a fact
that is lost when the coefficient of variation is calculated.

PSACNN results for a majority of the structures~(Fig.~\ref{fig:rd_threescanners}(a), Fig.~\ref{fig:rd_threescanners}(c), for example)
show lower median bias~($<2.5\%$), lower standard deviation, and fewer outliers. MALF has lower bias when segmenting HI~(Fig.\ref{fig:rd_threescanners}(g)), AM~(Fig.\ref{fig:rd_threescanners}(h)), whereas SAMSEG has better consistency
when segmenting the LV structure~(Fig.\ref{fig:rd_threescanners}(i)).
\textit{MEF} uses a different pulse sequence than all the other datasets
that have MPRAGE sequences. For most structures and algorithms shown in Fig.~\ref{fig:rd_threescanners}, we observe that \textit{MEF} shows the maximum relative volume difference from the mean (red boxplot in Figs.~\ref{fig:rd_threescanners}(a),(b) for example).
The Siemens acquisitions(\textit{TRIO} in blue and \textit{Siemens} in purple) show similar consistency to each other despite the differences in field strength~(3~T vs 1.5~T).

\subsubsection{Multi-TI Three Scanner Data}
\label{sec:uiodata}
In this experiment we tested the segmentation consistency of PSACNN on a more
extensive dataset collected at the University of Oslo. The dataset consists
of 24 subjects, each scanned on three Siemens scanners, Avanto, Skyra, and Prisma, with the MPRAGE sequence and two inversion times ($TI$)--850 ms and 1000 ms.
The dataset descriptions in detail are as follows:
\begin{itemize}
\item \textit{A850}: 3D MPRAGE, scanned on Siemens Avanto 1.5~T scanner, voxel-size $1.25\times1.25\times1.2$ mm$^3$, sagittal acquisition, $TR$=2400~ms, $TE$=3.61~ms, $TI$=850~ms, flip angle 8$^\circ$.
\item \textit{A1000}: 3D MPRAGE, scanned on Siemens Avanto 1.5~T scanner, voxel-size $1.25\times1.25\times1.2$ mm$^3$, sagittal acquisition, $TR$=2400~ms, $TE$=3.61~ms, $TI$=1000~ms, flip angle 8$^\circ$.
\item \textit{S850}: 3D MPRAGE, scanned on Siemens Skyra 3~T scanner, voxel-size $1\times1\times1$ mm$^3$, sagittal acquisition, $TR$=2300~ms, $TE$=2.98~ms, $TI$=850~ms, flip angle 8$^\circ$.
\item \textit{S1000}: 3D MPRAGE, scanned on Siemens Skyra 3~T scanner, voxel-size $1\times1\times1$ mm$^3$, sagittal acquisition, $TR$=2400~ms, $TE$=2.98~ms, $TI$=1000~ms, flip angle 8$^\circ$.
\item \textit{P850}: 3D MPRAGE, scanned on Siemens Prisma 3~T scanner, voxel-size $1\times1\times1$~mm$^3$, sagittal acquisition, $TR$=2400~ms, $TE$=2.22~ms, $TI$=850~ms, flip angle 8$^\circ$
\item \textit{P1000}: 3D MPRAGE, scanned on Siemens Prisma 3~T scanner, voxel-size $0.8\times0.8\times0.8$ mm$^3$, sagittal acquisition, $TR$=2400~ms, $TE$=2.22~ms, $TI$=1000~ms, flip angle 8$^\circ$
\end{itemize}

We calculate the coefficient of variation of structure volumes across
these six datasets similar to our experiment in Section~\ref{sec:fourscanner},
and are shown in Table~\ref{tab:coeffvar_sixscanners}.
\begin{table}[!t]
	\caption{Mean (and std. dev.) of the coefficient of variation (in $\%$) of
	structure volumes	over all six datasets with all algorithms.}
	\label{tab:coeffvar_sixscanners} \tabcolsep 0pt
	\vspace*{0.6em}
	\centerline{
	\resizebox{1\textwidth}{!} {
		\begin{tabular}{l c c c c c c c c c c c}
		  \toprule  & \hspace*{6.0ex} &
			\multicolumn{1}{c}{\textit{ASEG}} & \hspace*{4.0ex} &
			\multicolumn{1}{c}{\textit{SAMSEG}} & \hspace*{4.0ex} &
			\multicolumn{1}{c}{\textit{MALF}} & \hspace*{4.0ex} &
			\multicolumn{1}{c}{\textit{CNN}} & \hspace*{4.0ex} &
			\multicolumn{1}{c}{\textit{PSACNN}}
			\\
			 &&
			 \textbf{Mean} \textbf{~(Std)} &&
			 \textbf{Mean} \textbf{~(Std.)} &&
			 \textbf{Mean} \textbf{~(Std)} &&
			 \textbf{Mean} \textbf{~(Std.)} &&
			 \textbf{Mean} \textbf{~(Std.)} \\
			\cmidrule{3-3}
			\cmidrule{5-5} \cmidrule{7-7} \cmidrule{9-9} \cmidrule{11-11}
			 WM && \0{}2.82 (0.71) && \0{}3.06 (0.92) && \0{}2.30 (0.80) && \0{}6.01 (0.98) && \textbf{1.91} (0.70)$^*$ \\
			 CT && \0{}2.68 (0.67) && \textbf{1.42} (0.77)$^*$ && \0{}1.80 (0.82) && \0{}4.57 (1.37) && 1.85 (0.70)\0{} \\
			 TH && \0{}3.70 (1.32) && \0{}1.93 (0.72) && \0{}1.37 (0.59) && \0{}3.38 (1.35) && \textbf{0.85} (0.60)$^*$ \\
			 CA && \0{}2.02 (0.60) && \0{}2.38 (1.31) && \0{}1.80 (1.05) && \textbf{\0{}1.68} (1.33) && \textbf{1.68} (0.85)\0{} \\
			 PU && \0{}2.62 (1.22) && \0{}2.88 (1.78) && \textbf{\0{}1.24} (0.74) && \0{}3.08 (1.13) && 1.27 (0.86)\0{} \\
			 PA && \0{}4.67 (2.50) && \0{}2.10 (2.29) && \0{}2.65 (1.41) && \0{}5.69 (1.41) && \textbf{1.17} (0.56)$^*$ \\
			 HI && \0{}2.48 (0.78) && \0{}2.23 (3.54) && \0{}1.99 (1.60) && \0{}6.31 (2.08) && \textbf{1.00} (0.41)$^*$ \\
			 AM && \0{}6.50 (1.79) && \0{}2.44 (2.42) && \0{}1.93 (1.28) && \0{}3.87 (1.28) && \textbf{1.54} (0.56)\0{} \\
			 LV && \textbf{\0{}2.00} (0.82) && \0{}2.33 (1.36) && \0{}2.66 (2.38) && \0{}5.94 (1.87) && 2.21 (0.86)\0{} \\
			\bottomrule
		 \end{tabular}
		  }
			}
	\vspace{0.5em}
	\centerline{
		\scriptsize{ \begin{minipage}{0.93\textwidth} \textbf{Bold} results show
		the minimum coefficient of variation. * indicates significantly lower~($p < 0.05$, using the paired Wilcoxon signed rank test) coefficient of variation than the next best method. \end{minipage} } }
\end{table}
From Table~\ref{tab:coeffvar_sixscanners} we can observe that PSACNN structure
volumes have the lowest coefficient of variation for six of the nine structures.
For WM, TH, PA, and HI structures PSACNN coefficient of variation is significantly lower than the next best method~($p < 0.05$ using the paired Wilcoxon signed rank test). SAMSEG is significantly lower than MALF
for the CT structure. As compared to Table~\ref{tab:coeffvar_threescanners}
the coefficients of variation are lower as the input MPRAGE sequences
are the preferred input for all of these methods~(except SAMSEG).

\begin{sidewaysfigure}[!ht] \tabcolsep 0pt
\label{fig:six}
	\centerline{
		\begin{tabular}{ccc}
			\includegraphics[width=.33\textwidth]{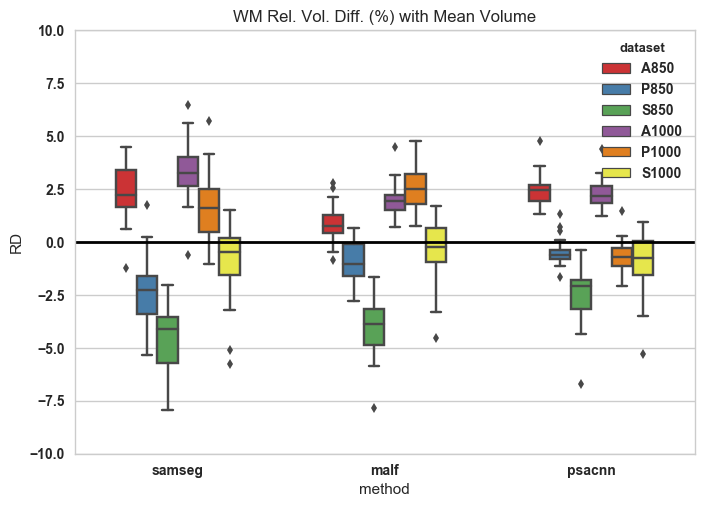} &
			\includegraphics[width=.33\textwidth]{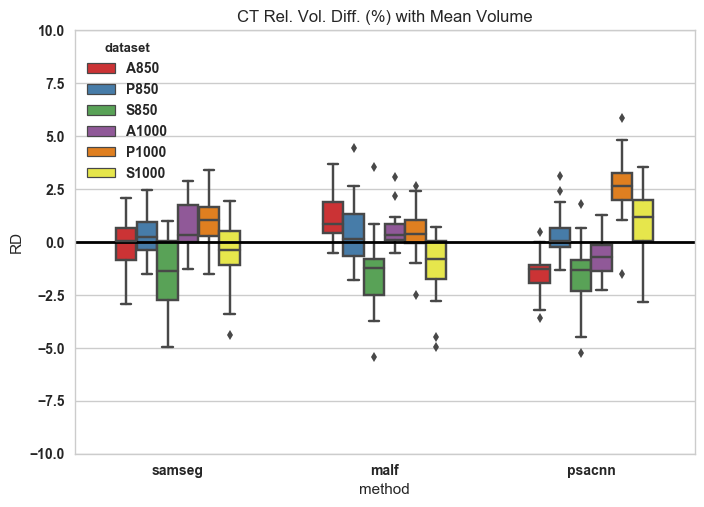} &
			\includegraphics[width=.33\textheight]{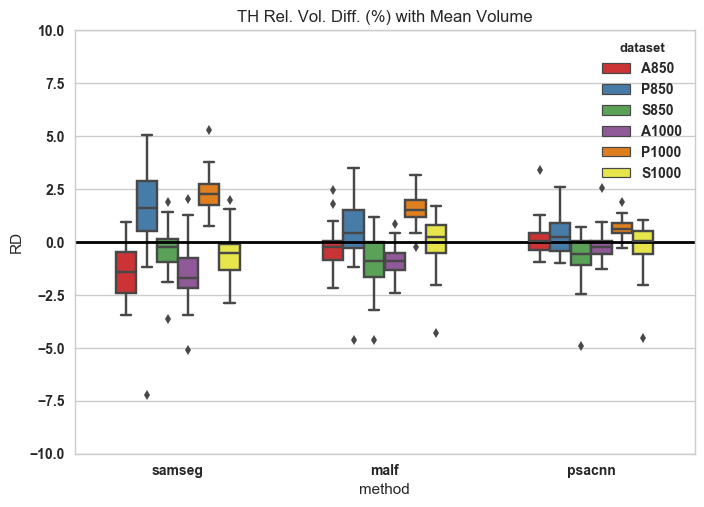}
			\\
			[-1.5em]
			{\textbf{(a)}}\hspace*{.25\textwidth} &
			{\textbf{(b)}}\hspace*{.25\textwidth} &
			{\textbf{(c)}}\hspace*{.25\textwidth}
			\\
	  	\includegraphics[width=.33\textwidth]{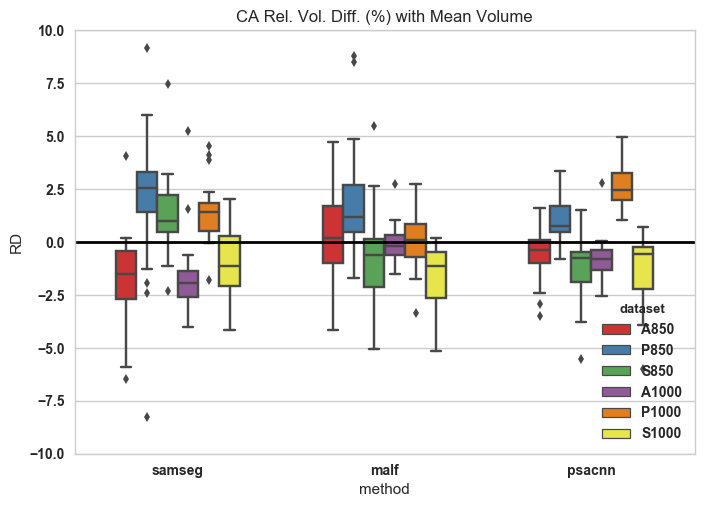} &
			\includegraphics[width=.33\textwidth]{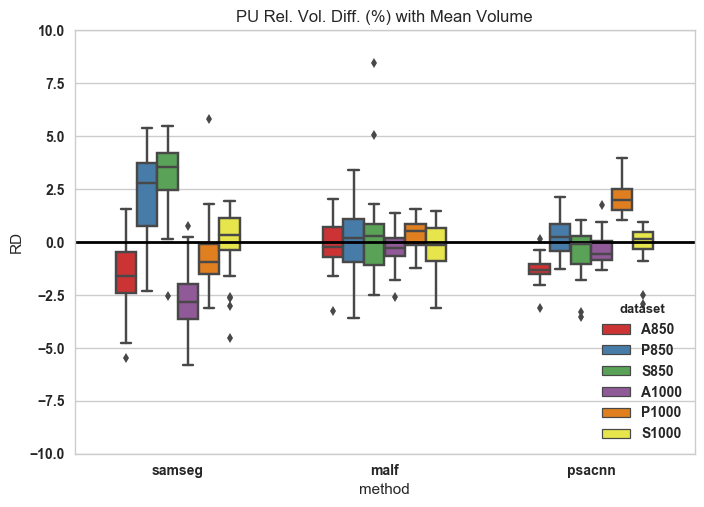} &
			\includegraphics[width=.33\textwidth]{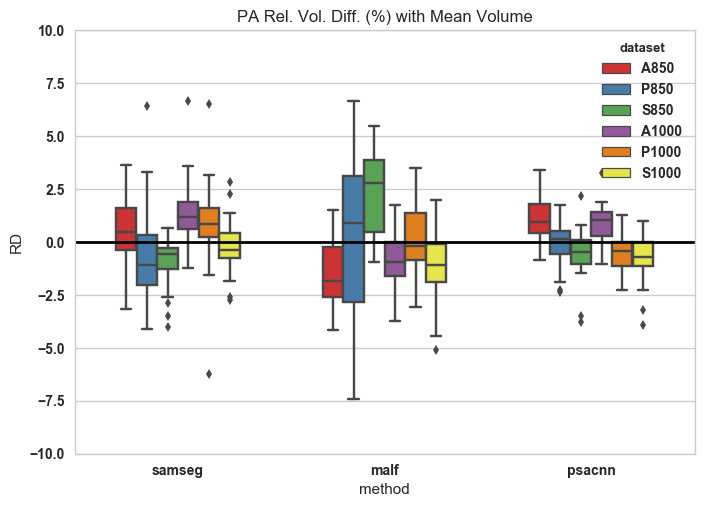} \\
			[-1.5em]
			{\textbf{(d)}}\hspace*{.25\textwidth} &
			{\textbf{(e)}}\hspace*{.25\textwidth} &
			{\textbf{(f)}}\hspace*{.25\textwidth}
			\\
			\includegraphics[width=.33\textwidth]{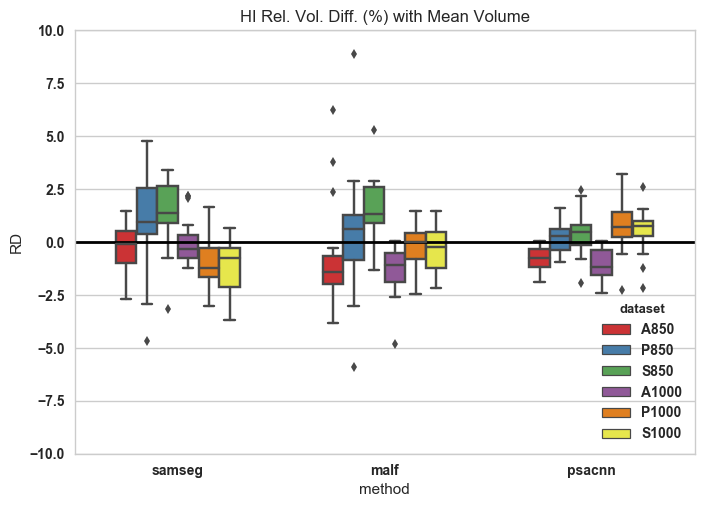} &
			\includegraphics[width=.33\textwidth]{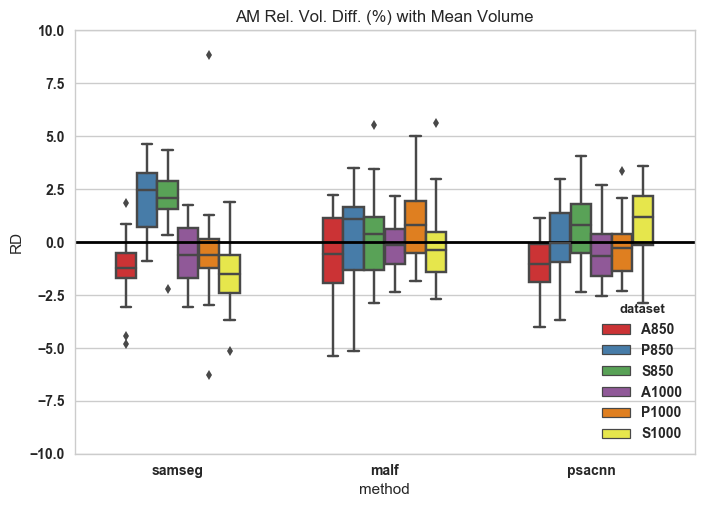} &
			\includegraphics[width=.33\textwidth]{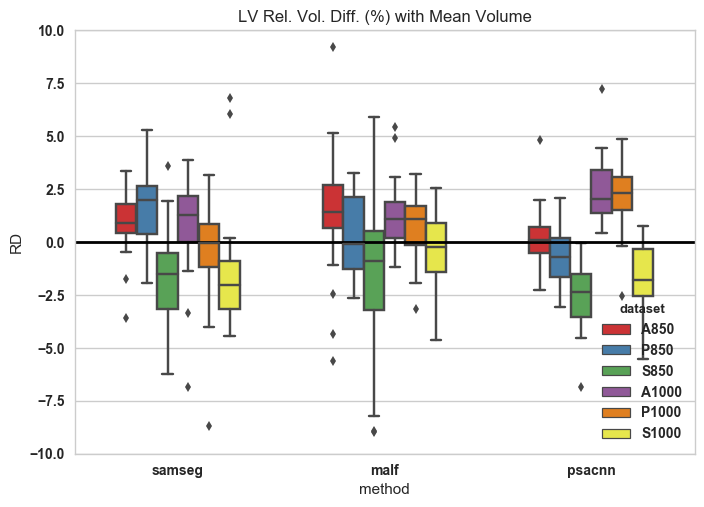}
			\\
			[-1.5em]
			{\textbf{(g)}}\hspace*{.25\textwidth} &
			{\textbf{(h)}}\hspace*{.25\textwidth} &
			{\textbf{(i)}}\hspace*{.25\textwidth}
	\end{tabular}
	}
	\caption{Signed relative difference of volumes of (a) WM, (b) CT, (c) TH, (d) CA, (e) PU, (f) PA, (g) HI, (h) AM, (i) LV from mean structure volume over all four datasets. In each figure, left group of boxplots shows SAMSEG, middle shows MALF, right-most shows PSACNN. The colors denote different datasets, \textit{A850}~(red), \textit{P850}~(blue), \textit{S850}~(green), \textit{A1000}~(purple),	\textit{P1000}~(orange), \textit{S1000}~(yellow). }
	\label{fig:rd_sixscanners}
\end{sidewaysfigure}

In Figs.~\ref{fig:rd_sixscanners}(a)--(i) we show the mean~(over 24 subjects)
signed relative volume difference~(in $\%$) with the mean structure volume
averaged over all six acquisition for nine structures. We focus on SAMSEG,
MALF, and PSACNN for this analysis. The boxplot colors denote different datasets, \textit{A850}~(red), \textit{P850}~(blue), \textit{S850}~(green), \textit{A1000}~(purple), \textit{P1000}~(orange), \textit{S1000}~(yellow). There is a small, statistically significant bias (difference from 0, thick black line) for most structures and algorithms. The bias is much smaller than that observed in Fig.~\ref{fig:rd_threescanners} for the Four Scanner Dataset.
PSACNN results for a majority of the structures~(for example, Figs.~\ref{fig:rd_sixscanners}(a), (c), (d), (f))
show lower median~($<2.5\%$), lower standard deviation, and fewer outliers. MALF has lower bias when segmenting, CT~(\ref{fig:rd_sixscanners}(b)), whereas SAMSEG has better consistency when segmenting the LV structure~(\ref{fig:rd_sixscanners}(i)).

From Fig.~\ref{fig:rd_sixscanners}(a), which shows the relative volume
difference for WM--which can act as an indicator of the gray-white contrast
present in the image--we can see that for all methods, images from same
scanner, regardless of TI, show similar relative volume differences.
The WM relative volume difference for \textit{A850} is most similar to \textit{A1000}, and similarly for the Prisma and Skyra datasets.
For PSACNN, this pattern exists for other structures such as CA, PA, PU, HI,
AM, LV as well. This suggests that change in the image contrast due to change in the inversion time $TI$ from 850~ms to 1000~ms, has a smaller effect on PSACNN segmentation consistency than a change of scanners (and field strengths).

\subsubsection{MPRAGE and T2-SPACE Segmentation}
\label{sec:fsmgreve}
\begin{figure}[!hp] \tabcolsep 1pt
	\centerline{
		\begin{tabular}{cccc}
			\includegraphics[width=.25\textwidth]{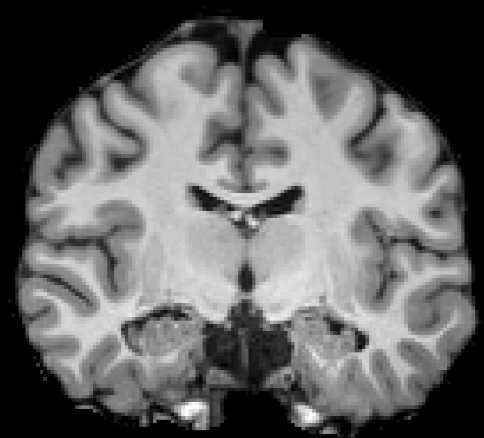} &
			\includegraphics[width=.25\textwidth]{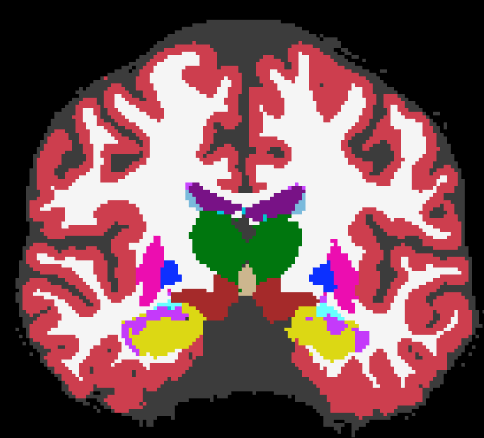} &
			\includegraphics[width=.25\textwidth]{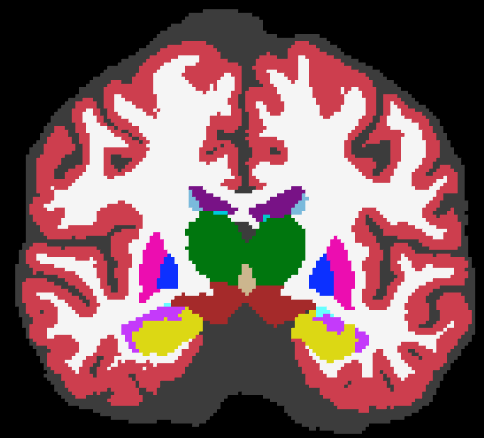} &
			\includegraphics[width=.25\textwidth]{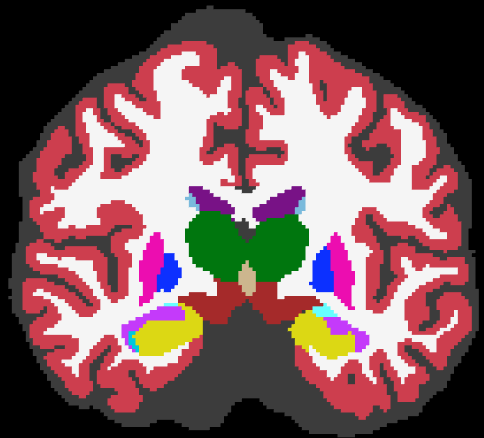}\\
			 (a) MPRAGE & (b) SAMSEG & (c) MALF & (d) PSACNN\\
			\includegraphics[width=.25\textwidth]{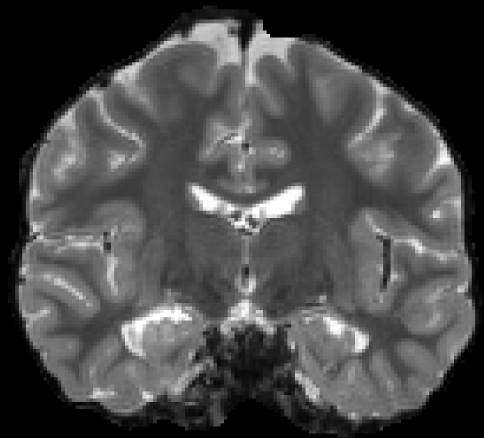}&
			\includegraphics[width=.25\textwidth]{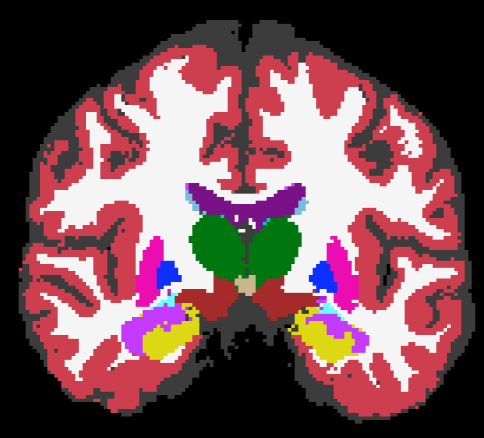}&
			\includegraphics[width=.25\textwidth]{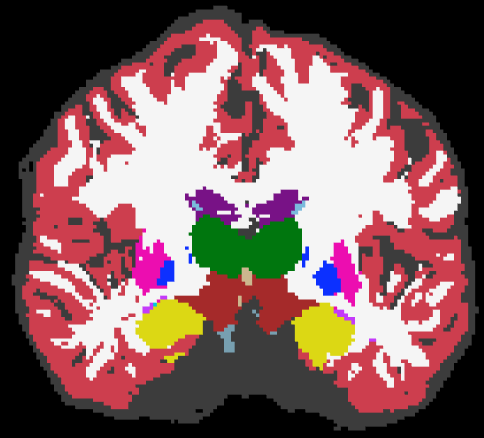}&
			\includegraphics[width=.25\textwidth]{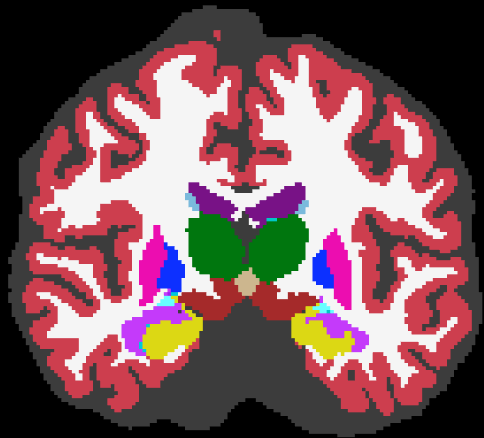}\\
			(e) T2-SPACE & (f) SAMSEG & (g) MALF & (h) PSACNN
	\end{tabular}
	}
	\caption{Input (a) MPRAGE and (e) T2-SPACE images with segmentations by SAMSEG (b) and (f), MALF (c) and (g), and PSACNN (d) and (h)}
	\label{fig:fsmgreve}
\end{figure}

In this experiment we use the same trained PSACNN network to segment
co-registered MPRAGE and T2-SPACE acquisitions for 10 subjects that
were acquired in the same imaging session. The acquisition
details for both the sequences are as follows:
\begin{itemize}
\item \textit{MPRAGE}: 3D MPRAGE, scanned on Siemens Prisma 3~T scanner, voxel-size $1\times1\times1$ mm$^3$, sagittal acquisition, $TR$=2530~ms, $TE$=1.69~ms, $TI$=1250~ms, flip angle 7$^\circ$.
\item \textit{T2-SPACE}: 3D T2-SPACE, scanned on Siemens Prisma 3~T scanner, voxel-size $1\times1\times1$ mm$^3$, sagittal acquisition, $TR$=3200~ms, $TE$=564~ms,  flip angle 120$^\circ$.
\end{itemize}
Figures~\ref{fig:fsmgreve}(a)--(i) show the input MPRAGE and T2-SPACE images
with their corresponding segmentations by SAMSEG, MALF, and PSACNN. ASEG
and CNN are not built or trained for non-$T_1$-w images so we did
not use them for comparison. We calculated the absolute relative volume difference of structures segmented for the \textit{T2-SPACE} dataset with
the same on the \textit{MPRAGE} dataset.
Table~\ref{tab:rd_fsmgreve} shows the absolute relative volume difference
for a subset of the structures. PSACNN shows the minimum absolute relative
volume difference for six of the nine structures, with significantly better
performance for subcortical structures including TH, CA, PU, and HI. SAMSEG
has significantly lower volume difference for PA and LV. All algorithms
shows a large volume difference for lateral ventricles~(LV). The LV has a complex geometry with a long boundary and a segmentation one-two voxels inside or outside
the LV of the \textit{MPRAGE} image can cause a large relative volume
difference. Synthetic T2-SPACE images generated in the PSACNN  augmentation can match the overall image contrast with the real T2-SPACE images, however we believe the intensity gradient from white matter~(dark) to CSF~(very bright) is not accurately matched to the intensity gradient in the real, test T2-SPACE images, possibly due to resolution differences in the training and test data. This points to avenues of potential improvement in PSACNN augmentation.
MALF~(Fig.~\ref{fig:fsmgreve}(c) and Fig.~\ref{fig:fsmgreve}(g)) shows a diminished performance despite using a cross-modal registration metric like cross-correlation in the ANTS registration between the $T_1$-w atlas images and the $T_2$-w target images.

Since these two acquisitions were co-registered, we can calculate the Dice
overlap between MPRAGE and T2-SPACE labeled structures, as shown in Table~\ref{tab:dice_fsmgreve}. As we can observe, PSACNN shows a significantly
higher overlap than the next best method on six of the nine structures, and
tying with SAMSEG for thalamus~(TH). SAMSEG shows a significantly
higher overlap for PA and LV.
\begin{table}[!t]
	\caption{Mean (and std. dev.) of the absolute relative volume difference (in $\%$) of
	structure volumes	for \textit{T2-SPACE} dataset with respect to \textit{MPRAGE} dataset.}
	\label{tab:rd_fsmgreve} \tabcolsep 0pt
	\vspace*{0.6em}
	\centerline{
			\begin{tabular}{l c c c c c c c }
		  \toprule  & \hspace*{6.0ex} &
			\multicolumn{1}{c}{\textit{SAMSEG}} & \hspace*{4.0ex} &
			\multicolumn{1}{c}{\textit{MALF}} & \hspace*{4.0ex} &
			\multicolumn{1}{c}{\textit{PSACNN}} & \hspace*{4.0ex}
   		\\
			 &&
			 \textbf{Mean} \textbf{~(Std)} &&
			 \textbf{Mean} \textbf{~(Std.)} &&
			 \textbf{Mean} \textbf{~(Std)}
			  \\
				\cmidrule{3-3}
			\cmidrule{5-5} \cmidrule{7-7}
			 WM && \0{}5.20 (2.38) && \0{}5.01 (2.74) && \textbf{3.74} (1.30)\0{}  \\
			 CT && \0{}8.22 (1.04) && \0{}2.98 (2.16) && \textbf{2.95} (1.42)\0{} \\
			 TH && \0{}7.71 (2.72) && 13.60 (6.13) && \textbf{4.22} (2.17)$^*$  \\
			 CA && \0{}9.15 (2.66) && 34.51 (13.23) && \textbf{3.98} (2.72)$^*$  \\
			 PU && \0{}5.31 (1.99) && 10.60 (7.92) && \textbf{2.48} (1.57)$^*$  \\
			 PA && \textbf{2.85} (1.70)$^*$ && 42.36 (15.01) && 13.35 (10.12)  \\
			 HI && 14.17 (5.82) && 30.12 (12.14) && \textbf{3.18} (2.05)$^*$  \\
			 AM && 12.46 (4.50) && \textbf{10.49} (4.48) && 10.69 (4.42) \\
			 LV && \textbf{16.44} (5.93)$^*$ && 41.92 (47.89) && 25.83 (7.13)  \\
			\bottomrule
		 \end{tabular}
				}
	\vspace{0.5em}
	\centerline{
		\scriptsize{ \begin{minipage}{0.93\textwidth} \textbf{Bold} results show
		the minimum absolute relative difference. * indicates significantly lower~($p < 0.05$, using the paired Wilcoxon signed rank test) absolute relative volume difference than the next best method. \end{minipage} } }
\end{table}

\begin{table}[!t]
	\caption{Mean (and std. dev.) of the Dice coefficient for	structures of \textit{T2-SPACE} dataset with respect to \textit{MPRAGE} dataset.}
	\label{tab:dice_fsmgreve} \tabcolsep 0pt
	\vspace*{0.6em}
	\centerline{
			\begin{tabular}{l c c c c c c c }
		  \toprule  & \hspace*{6.0ex} &
			\multicolumn{1}{c}{\textit{SAMSEG}} & \hspace*{4.0ex} &
			\multicolumn{1}{c}{\textit{MALF}} & \hspace*{4.0ex} &
			\multicolumn{1}{c}{\textit{PSACNN}} & \hspace*{4.0ex}
   		\\
			 &&
			 \textbf{Mean} \textbf{~(Std)} &&
			 \textbf{Mean} \textbf{~(Std.)} &&
			 \textbf{Mean} \textbf{~(Std)}
			  \\
			\cmidrule{3-3}
			\cmidrule{5-5} \cmidrule{7-7}
			 WM && \0{}0.91 (0.01) && \0{}0.57 (0.01) && \textbf{0.94} (0.00)$^*$  \\
			 CT && \0{}0.85 (0.01) && \0{}0.59 (0.00) && \textbf{0.89} (0.01)$^*$ \\
			 TH && \0{}0.92 (0.01) && \0{}0.80 (0.03) && \textbf{0.92} (0.01)\0{}  \\
			 CA && \0{}0.88 (0.01) && \0{}0.65 (0.04) && \textbf{0.90} (0.00)$^*$  \\
			 PU && \0{}0.88 (0.02) && \0{}0.71 (0.05) && \textbf{0.91} (0.02)$^*$  \\
			 PA && \textbf{0.89} (0.02)$^*$ && \0{}0.71 (0.03) && 0.81 (0.05)\0{}  \\
			 HI && \0{}0.84 (0.04) && \0{}0.60 (0.03) && \textbf{0.88} (0.02)$^*$  \\
			 AM && \0{}0.83 (0.03) && \0{}0.65 (0.06) && \textbf{0.88} (0.01)$^*$ \\
			 LV && \textbf{0.90} (0.03)$^*$ && \0{}0.63 (0.10) && 0.87 (0.03)\0{}  \\
			\bottomrule
		 \end{tabular}
				}
	\vspace{0.5em}
	\centerline{
		\scriptsize{ \begin{minipage}{0.93\textwidth} \textbf{Bold} results show
		the maximum Dice coefficient. * indicates significantly higher~($p < 0.05$, using the paired Wilcoxon signed rank test) Dice than the next best method. \end{minipage} } }
\end{table}

%
%
%

\section{Conclusion and Discussion}
\label{sec:conclusion}
We have described PSACNN, which uses a new strategy for training CNNs for the
purpose of consistent whole brain segmentation across multi-scanner and multi-sequence MRI data. PSACNN shows significant improvement over state-of-the-art segmentation algorithms in terms accuracy based on
our experiments on multiple manually labeled datasets acquired with different
acquisition settings as shown in Section~\ref{sec:accuracy}.

For modern large, multi-scanner datasets, it is imperative for segmentation
algorithms to provide consistent outputs on a variety of differing acquisitions.
This is especially important when imaging is used to quantify the efficacy
of a drug in Phase III of its clinical trials. Phase III trials are required
to be carried out in hundreds of centers with access to a large subject
population across tens of countries~\citep{murphy2010imaging}, where
accurate and consistent segmentation across sites is critical.
The pulse sequence parameter estimation-based augmentation strategy in training
PSACNN allows us to train the CNN for a wide range of input pulse sequence
parameters, leading to a CNN that is robust to input variations. In Section~\ref{sec:consistency}, we compared the variation in structure volumes
obtained via five segmentation algorithms for a variety of $T_1$-weighted~(and $T_2$-weighted) inputs and
PSACNN showed the highest consistency among all for many of the structures,
with many structures showing less than $1\%$ coefficient of variation. Such
consistency enables detection of subtle changes in structure volumes in cross-sectional~(healthy vs. diseased) and longitudinal studies, where
variation can occur due to scanner and protocol upgrades.

PSACNN training includes augmented, synthetic MRI images generated by approximate forward models of the pulse sequences. Thus, as long as we can
model the  pulse sequence imaging equation, we can include multiple pulse
sequence versions of the same anatomy in the augmentation and rely on the
representative capacity of the underlying CNN architecture~(U-Net in our case)
to learn to map all of them to the same label map in the training data. We
formulated approximate forward models of pulse sequences to train
a network that can simultaneously segment MPRAGE, FLASH, SPGR, and T2-SPACE images consistently. Moreover, the NMR maps used in PSACNN need not perfectly match maps acquired using relaxometry sequences. They merely act as parameters
that when used in our approximate imaging equation accurately generate the
observed image intensities.

Being a CNN-based method, PSACNN is 2--3 orders of magnitude faster than current
state-of-the-art methods such as MALF and SAMSEG. On a system with an NVidia
GPU~(we tested Tesla K40, Quadro P6000, P100, V100) PSACNN completes a segmentation for a single volume within 45 seconds.  On a single thread CPU,
it takes about 5 minutes. SAMSEG and MALF take 15--30~minutes if they are heavily multi-threaded. On a single thread SAMSEG takes about 60~minutes, whereas MALF can take up to 20 hours. Pre-processing for PSACNN
(inhomogeneity correction, skullstripping) can take up to 10---15 minutes, whereas SAMSEG does not need any pre-processing.
With availability of training NMR maps of whole head images, PSACNN will be able to drop the time consuming skullstripping step. A fast, consistent segmentation method like PSACNN considerably speeds up neuroimaging pipelines that traditionally take hours to completely process a single subject.

PSACNN in its present form has a number of limitations we intend to work
on in the future. The current augmentation does not include matching resolutions
of the training and test data. This is the likely reason for overestimation
of lateral ventricles in the T2-SPACE segmentation, as the training data
and the synthetic T2-SPACE images generated from them have a slightly lower resolution than the test T2-SPACE data. Acquiring higher resolution training dataset with relaxometry
sequences for ground truth $P_D$, $T_1$, and $T_2$ values will also help.
Our pulse sequence models are approximate and can be robustly estimated. However, we do not account for any errors in the pulse sequence parameter
estimation and the average NMR parameters. A probabilistic model that models
these uncertainties to further make the estimation robust is a future goal.
The pulse sequence parameter estimation also assumes the same set of parameters
across the whole brain. It does not account for local differences, for example
variation in the flip angle with B1 inhomogeneities.

PSACNN currently uses $96\times96\times96$-sized patches to learn and
predict the segmentations. This is a forced choice due to limitations in the
GPU memory. However, using entire images would require significantly more training data than $20$--$40$ labeled subjects. It is unknown how such a network would handle unseen anatomy. Further, it will be interesting to extend PSACNN to handle pathological cases involving tumors and lesions.

In conclusion, we have described PSACNN, a fast, pulse sequence resilient
whole brain segmentation approach. The code will be made available as a part of the FreeSurfer development version repository~(\url{https://github.com/freesurfer/freesurfer}). Further validation and testing will need to be carried out before its release.

\section*{Acknowledgments}
This research was carried out in whole or in part at the Athinoula A. Martinos Center for Biomedical Imaging at the Massachusetts General Hospital, using resources provided by NIH grant 5U24NS100591-02 and the Center for Functional Neuroimaging Technologies, P41EB015896, a P41 Biotechnology Resource Grant supported by the National Institute of Biomedical Imaging and Bioengineering (NIBIB), National Institutes of Health, and 1R01EB023281-01
FreeSurfer Development, Maintenance, and Hardening, and in part through the BRAIN Initiative Cell Census Network grant U01MH117023.
This project has also received funding from the European Union's Horizon 2020
research and innovation programme under the Marie Skłodowska-Curie grant
agreement No 765148, Danish Council for Independent Research under grant
number DFF-6111-00291, and NIH: R21AG050122 National Institute on Aging.

\bibliographystyle{model2-names}
\bibliography{ms}

\end{document}